\documentclass[11pt]{article}

\usepackage[preprint]{acl}
\usepackage{times}
 \usepackage{booktabs,multirow}
\usepackage{latexsym}

\usepackage[T1]{fontenc}

\usepackage[utf8]{inputenc}

\usepackage{microtype}

\usepackage{inconsolata}

\usepackage{graphicx}
\usepackage{amsmath}
\usepackage{amssymb}
\usepackage{booktabs}
\usepackage{xcolor}
\usepackage[table]{xcolor}

\usepackage{pifont}

\newcommand{\cmark}{\textcolor{green!60!black}{\ding{51}}}
\newcommand{\xmark}{\textcolor{red!70!black}{\ding{55}}}
\title{TVWorld: Foundations for Remote-Control TV Agents}

\author{ Zhantao Ma$^{1}$\footnotemark[1], Quanfeng Lu$^{1}$\thanks{Equal Contribution}, Shuai Zhong$^{1}$, Dahai Yu$^{3}$ \\\textbf{Ping Luo}$^{1}$, \textbf{Michael K. Ng}$^{2}$\thanks{Corresponding Author: michael-ng@hkbu.edu.hk} \\ 
$^{1}$The University of Hong Kong \quad$^{2}$Hong Kong Baptist University \\
$^{3}$TCL Corporate Research (Hong Kong) Co., Ltd\\
\url{https://github.com/Lqf-HFNJU/TVTheseus}}

\begin{document}
\maketitle
\begin{abstract}
Recent large vision–language models (LVLMs) have demonstrated strong potential for device control. However, existing research has primarily focused on point-and-click (PnC) interaction, while remote-control (RC) interaction commonly encountered in everyday TV usage remains largely underexplored. To fill this gap, we introduce \textbf{TVWorld}, an offline graph-based abstraction of real-world TV navigation that enables reproducible and deployment-free evaluation. On this basis, we derive two complementary benchmarks that comprehensively assess TV-use capabilities: \textbf{TVWorld-N} for topology-aware navigation and \textbf{TVWorld-G} for focus-aware grounding.
These benchmarks expose a key limitation of existing agents: insufficient topology awareness for focus-based, long-horizon TV navigation. Motivated by this finding, we propose a \emph{Topology-Aware Training} framework that injects topology awareness into LVLMs. Using this framework, we develop \textbf{TVTheseus}, a foundation model specialized for TV navigation. TVTheseus achieves a success rate of $68.3\%$ on TVWorld-N, surpassing strong closed-source baselines such as Gemini 3 Flash and establishing state-of-the-art (SOTA) performance. Additional analyses further provide valuable insights into the development of effective TV-use agents.
\end{abstract}

\begin{figure*}[h!]
    \centering
    \includegraphics[width=0.85\textwidth]{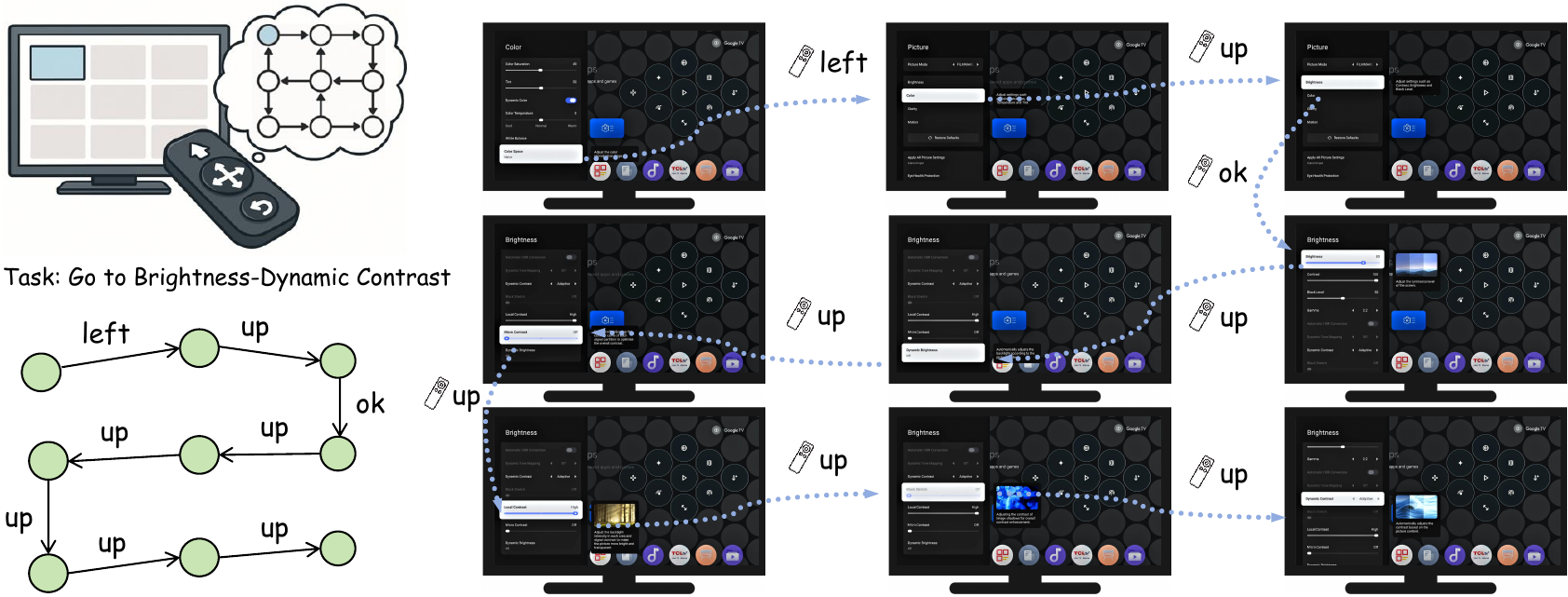}
    \caption{Illustration of focus-based remote-control TV interaction: discrete key presses (e.g. \texttt{LEFT/UP/OK}) move a highlight across UI elements, inducing UI state transitions toward the target. This process can be formulated as path finding on a topology graph whose nodes are UI states and edges correspond to key-induced transitions.}
    \label{fig:TVexample}
\end{figure*}

\section{Introduction}
When discussing how Large Vision–Language Models (LVLMs) operate in user interface (UI) environments, existing work predominantly assumes a point-and-click interaction paradigm~\citep{wang2025uitars2, wang2025opencua, ye2025mobile}, where a cursor or fingertip directly selects on-screen targets. However, this assumption does not generalize to smart televisions (TVs), a widely deployed and increasingly common media platform~\citep{strategy-analytics-2021-smart-tv-forecast-pressrelease}, where interaction is mediated through remote-control navigation~\citep{hong-rivoal-2019-css-nav-1} rather than direct pointing. 
TV interfaces are inherently focus-based: navigation is performed by pressing directional keys on the remote control (e.g., the \texttt{UP} button), which move a discrete highlight across UI elements, and actions are executed only on the currently focused item (Fig.~\ref{fig:TVexample}).

This interaction paradigm gives rise to requirements that differ fundamentally from pointer-based UI controls. Effective TV navigation hinges on \emph{focus awareness}, which involves localizing the currently highlighted element within the global screen layout rather than detecting clickable elements in isolation, and \emph{topology-aware planning}, which entails perceiving the underlying UI topology and navigating through discrete, button-driven focus transitions over multiple steps. Yet, existing GUI control benchmarks for LVLM agents remain largely dominated by pointer primitives, leaving evaluation settings that capture the demands of TV navigation scarce.

To fill this gap, we introduce \textbf{TVWorld}, an offline graph-based abstraction of real-world TV navigation. TV interaction is driven by a small set of discrete remote-control keys and spans a limited set of stable UI screens, which makes a graph formulation natural: each UI state corresponds to a node, and each key press induces a transition recorded as a directed edge, enabling TV navigation to be captured as a finite graph. Based on this abstraction, we systematically traverse real TV interfaces via remote-control interaction to construct high-fidelity navigation graphs. Building on TVWorld, we derive two complementary benchmarks tailored to TV interaction. 
\textbf{TVWorld-N} is an offline interactive TV navigation environment for evaluating agents’ \emph{topology-aware planning} under focus-based remote-control, supporting both textual and visual goals. Operating purely on static graph assets, it is fully replayable and deployment-free (e.g., no VMs/emulators), and enables millisecond-level interaction, avoiding the instability and overhead of online GUI benchmarks~\cite{xie2024osworld, rawles2024}.
Complementarily, \textbf{TVWorld-G} evaluates \emph{focus-aware grounding} by requiring the agent to localize the currently highlighted element within the global screen layout using bounding-box annotations, directly reflecting the focus-based nature of TV control.

TVWorld further exposes a critical limitation of recent GUI agents: while they excel at pointer-based interaction, they struggle with the focus-based, long-horizon navigation required by TV control. We attribute this failure to the lack of \emph{topology awareness}, namely the ability to perceive focus structure and navigate through UI state transitions. To address this bottleneck, we design three types of topology-driven training traces that explicitly target fundamental behaviors in TV navigation, including goal-directed focus transitions, recovery from detours, and escape from stalled states. These traces form the basis of a two-stage \emph{Topology-Aware Training} framework that systematically builds focus awareness and topology-aware planning. Using this training strategy, we develop \textbf{TVTheseus}, a foundation model specialized for TV navigation. Comprehensive experiments demonstrate the effectiveness and robustness of our approach: TVTheseus achieves a success rate of $68.3$ on the out-of-domain TVWorld-N benchmark, outperforming strong closed-source baselines such as Gemini 3 Flash and GPT-5 mini, and attains SOTA performance on TVWorld-G with an accuracy of $81.8$ despite receiving no grounding-specific supervision, reflecting strong topology-aware planning and focus-awareness capabilities.

The contributions of this work are three-fold:
1) we introduce \textbf{TVWorld}, a static benchmark for interactive TV navigation, together with two complementary evaluation suites;
2) we propose a \emph{Topology-Aware Training} framework tailored to focus-based TV interaction; and
3) using this framework, we develop \textbf{TVTheseus}, a foundation model for TV navigation, whose effectiveness is validated through extensive empirical evaluation.

\section{TVWorld}\label{sec:tvworld}
This section introduces \textbf{TVWorld}, which transforms real-world remote-control TV interaction into an offline asset for LVLM-based agent navigation. We first formulate TV navigation as a graph search over UI states and define TV-specific tasks (Sec.~\ref{sec:graph}). We then describe our on-device data collection pipeline for graph construction (Sec.~\ref{subsec:tv-graph}). Based on the resulting graphs, we introduce two benchmarks: \textbf{TVWorld-N} for end-to-end topology-aware navigation (Sec.~\ref{subsec:offline-simulated-device-dataset}) and \textbf{TVWorld-G} for focus-aware grounding evaluation (Sec.~\ref{subsec:Focus-Aware}).

\subsection{Task Formulation}
\label{sec:graph}
Remote-control navigation can be modeled as a discrete state--action trajectory. Actions come from a small, fixed set of keys, and a state is represented by a screenshot together with its visible focus highlight. After each key press, the UI renders a new screen: sometimes the focus shifts within the same page, and sometimes the interface switches to another page. This “one key, one transition” view is naturally captured by an action-labeled directed multigraph, where nodes are UI states and edges are key-triggered state changes.

\paragraph{TV Navigation Graph.}
We define the \emph{TV Navigation Graph} as \(\mathcal{G}=(\mathcal{V},\mathcal{E},\lambda)\) with \(\mathcal{E}\subseteq \mathcal{V}\times \mathcal{A}\times \mathcal{V}\). Here \(\mathcal{V}\) contains all UI states reachable from an anchor screen (e.g., Home) by feasible key sequences; each \(u\in\mathcal{V}\) is a UI state that contains one screenshot with its focus highlight. \(\mathcal{A}\) is the set of atomic remote actions. An edge \((u,a,v)\in\mathcal{E}\) means that pressing \(a\in\mathcal{A}\) at state \(u\) yields the next state \(v\). We write the transition as \(T:\mathcal{V}\times\mathcal{A}\to\mathcal{V}\) with \(T(u,a)=v\) whenever \((u,a,v)\in\mathcal{E}\). Each node carries a label \(\lambda:\mathcal{V}\to\mathcal{L}\) given by \(\lambda(u)=(S(u),\,\mathcal{A}(u),\,m(u))\), where \(S(u)\) is the screenshot, \(\mathcal{A}(u)\subseteq\mathcal{A}\) lists valid actions at \(u\), and \(m(u)\) is optional metadata (e.g., text cues or a view-tree).

\paragraph{Action Set.} The action set of TVWorld comprises $9$ kinds of actions: \texttt{UP}, \texttt{DOWN}, \texttt{LEFT}, \texttt{RIGHT}, \texttt{EXIT}, \texttt{OK}, \texttt{HOME}, \texttt{SETTING}, and \texttt{FINISH}. The functionalities of these actions are summarized in Appendix~\ref{app:details of action_set}.

\paragraph{Topology-Aware Navigation.} TV topology-aware navigation can be modeled as a partially observable Markov decision process (POMDP)
$(\mathcal{S}, \mathcal{O}, \mathcal{A}, \mathcal{T})$, where $\mathcal{S}$ denotes the TV environment states,
$\mathcal{O}$ denotes observations (e.g., screenshots and textual cues), $\mathcal{A}$ is the set of remote-control actions
(see Appendix~\ref{app:details of action_set}), and $\mathcal{T}$ is the state transition.
At timestep $t$, the agent receives an observation $o_t \in \mathcal{O}$ composed of a task instruction $I$ (visual or textual),
the current TV screenshot, and optionally a history of previous observations. Based on $o_t$, the agent selects a remote-control key action $a_t \in \mathcal{A}$, such as \texttt{UP} or \texttt{DOWN}.
Executing $a_t$ updates the environment to a new state $s_{t+1} \in \mathcal{S}$ and produces a subsequent observation
$o_{t+1} \in \mathcal{O}$ (e.g., a refreshed screenshot). This agent--TV interaction proceeds until the agent emits the
terminal action \texttt{FINISH} or a predefined step budget is exhausted.

\paragraph{Focus-Aware Grounding.} Focus-aware grounding aims to localize the \emph{currently focused (highlighted)} UI element on a TV interface.
Formally, given a TV GUI screenshot $S$ and an instruction $I$, an agent $\pi$ predicts
the focused element's location as a bounding box $b=(x_1,y_1,x_2,y_2)\sim\pi(S,I),$ where $(x_1,y_1)$ and $(x_2,y_2)$ denote the top-left and bottom-right coordinates of the focused element.

\begin{figure}[h]
    \centering
    \includegraphics[width=1\linewidth]{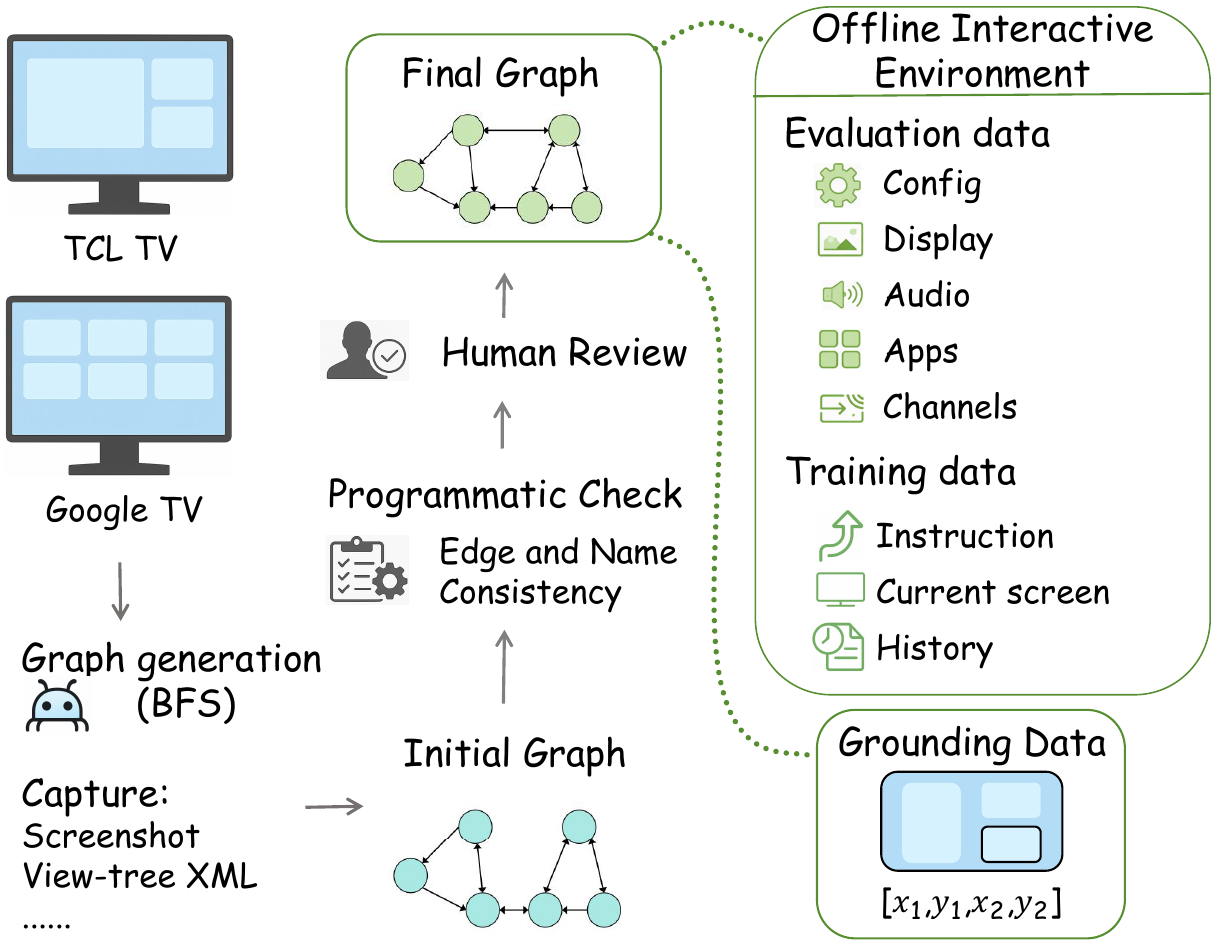}
    \caption{Overview of the TVWorld graphs collection pipeline. We perform BFS exploration on physical TV devices (TCL TV and Google TV) to construct initial UI-state graphs, while recording screenshots and view-tree metadata. Graphs are then refined through automated consistency checks and human inspection, producing finalized graphs together with the offline interactive environment for evaluation/training and grounding data.}
    \label{fig:placeholder}
\end{figure}
\subsection{TVWorld Toolkit}
\label{subsec:tv-graph}
We deployed an automated UI data collection system on physical smart TVs using an MT9655-based 4K platform, and adapted it to two product lines: \emph{TCL TV}\footnote{\href{https://www.tcl.com/global/en/qd-mini-led-tv}{TCL
 QD-Mini LED TV}} and \emph{Google TV}\footnote{\href{https://www.tcl.com/ca/en/products/home-theater/4-series/55-class-4-series-4k-uhd-led-smart-google-tv-55s446-ca}{4K
 Ultra HD Smart Google TV}}.  We designate the TCL TV platform for training, while reserving the Google TV platform exclusively for evaluation. The substantial differences in UI styles, layouts, and design elements between the two platforms enable a rigorous assessment of the agent's ability to generalize to out-of-domain UI environments; examples are provided in Appendix~\ref{appsec:Comparison of TV UI Layouts}. The data acquisition client communicates with the device via Android Debug Bridge (ADB) and a capture card to obtain UI hierarchy snapshots, focus metadata, and screenshots, which are logged per session for traceability. The system interfaces with the television through a hardware-level serial remote-control interface, which injects directional and other key inputs to induce UI state transitions under real-world interaction conditions. Through this combined software–hardware architecture, the system enables end-to-end capture of interaction trajectories and user interface representations.

\paragraph{Graph Collection Pipeline.} We build directed TV navigation graphs via BFS exploration starting from the homepage. Each node corresponds to a UI state, identified by the screen together with its focused element; the node also stores structured metadata and the associated view-tree dump. For every node, we expand the graph using a fixed, ordered sequence of remote-control key events. When a keypress changes the state, we either create a new node or match it to an existing one, and then add a directed edge labeled with that key. 
The crawl explicitly avoids a small set of sensitive entry points (e.g., factory reset and language switching), which are masked for safety. After collection, professional TV engineers assign standardized, unambiguous names to nodes so that labels align with page semantics and avoid duplicates.

\paragraph{Graph Quality Assurance.}
Each generated navigation graph undergoes a structured quality assurance process that integrates automated validation with expert review. Specifically, we perform (i) transition integrity checks, which identify missing inverse links (e.g., an \texttt{UP} move from node $u$ to node $v$ without a corresponding \texttt{DOWN}). (ii) naming and hierarchy verification, which flags duplicated node identifiers and inconsistencies in navigational relations (e.g., nodes connected via \texttt{LEFT} or \texttt{EXIT} whose names do not indicate the expected sub-level relation). (iii) human-in-the-loop validation, where all automatically flagged issues are examined and resolved by TV engineers who confirm node definitions, edge semantics, and local navigation behavior. Following these corrections, a senior engineer conducts an end-to-end audit of the full graph to confirm global consistency in topology, naming hierarchy, and directional behavior, and to ensure that the final graph is coherent and reliable.

\paragraph{TVWorld Statistics.} Using the proposed graph collection pipeline, we construct $6$ \emph{directed} TV navigation graphs spanning $5$ UI scenarios: \textsc{Config} (system-level configuration and global settings), \textsc{Display} (display and rendering), \textsc{Audio} (audio-related capabilities), \textsc{Apps} (app entry points and privacy), and \textsc{Channels} (hardware-interface components). All graphs are \emph{strongly connected}, i.e., for any pair of nodes, there exists a directed path between them. Beyond graph topology, each node is associated with rich UI metadata, including screenshots, view-tree structures, focus-related information, and other attributes. 
For data splits, we use the graph collected from TCL TV for training, and reserve the remaining $5$ graphs collected from Google TV for comprehensive evaluation. Detailed per-graph statistics are reported in Table~\ref{tab:tvworld_stats}.

\begin{table}[h]
\centering
\small
\caption{Statistics of TVWorld.}
\scalebox{1}{
\begin{tabular}{lcccc}
\toprule
\textbf{Platform} & \textbf{Scenario} & \textbf{Nodes} & \textbf{Edges} \\
\midrule
\rowcolor{gray!15}
\multicolumn{4}{l}{\textit{Train}} \\
 TCL TV    & Config   & 282 & 1{,}508 \\
\midrule
\rowcolor{gray!15}
\multicolumn{4}{l}{\textit{Test}} \\
Google TV & Config   &  169 & 878 \\
Google TV &  Display  & 62 & 276 \\
Google TV &  Audio    & 24 & 104 \\
Google TV &  Apps     & 33 & 147 \\
Google TV & Channels &  32 & 145 \\
\bottomrule
\end{tabular}}

\label{tab:tvworld_stats}
\end{table}

\begin{table*}[t]
\centering
\small
\caption{Comparison of mainstream GUI evaluation benchmarks.}

\scalebox{0.73}{\begin{tabular}{l c c c c c c c c}
\toprule
\textbf{Benchmark} &
\textbf{Platform} &
\textbf{Interactive Env.} &
\textbf{Deploy-Free} &
\textbf{Replayable} &
\textbf{\#Tasks} &
\textbf{Real-World} &
\textbf{Metadata} &
\textbf{Easy Task Gen.} \\

\midrule
AitW~\citep{zhang2024android} & Mobile  & \xmark & \cmark & \cmark & 2346 & \cmark & \xmark & \xmark \\
AndroidControl\cite{li2024effects} & Mobile  & \xmark & \cmark & \cmark & 15283 & \cmark & \cmark & \xmark \\
GUIOdyssey\cite{lu2025guiodyssey} & Mobile  & \xmark & \cmark & \cmark & 8334 & \cmark & \xmark & \xmark \\
MiniWoB++~\citep{liu2018reinforcement} & Web & \cmark & \xmark & \cmark & 114 & \xmark & \cmark & \xmark \\
WebArena~\citep{zhou2023webarena} & Web & \cmark & \xmark & \cmark & 812 & \cmark & \cmark & \xmark \\

OSWorld~\citep{xie2024osworld} & Desktop & \cmark & \xmark & \xmark & 369 & \cmark & \cmark & \xmark \\
WindowsAgentArena\citep{bonatti2024windows} & Desktop & \cmark & \xmark & \xmark & 154 & \cmark & \cmark & \xmark \\

Online-Mind2Web~\citep{xue2025illusion}&Web& \cmark& \xmark & \xmark& 300& \cmark& \cmark& \xmark\\
AndroidWorld~\citep{rawles2024} & Mobile & 
\cmark & \xmark & \xmark & 116 & \cmark & \cmark & \xmark \\
\midrule
\textbf{TVWorld-N (Ours)} & TV & \cmark & \cmark & \cmark & 500 & \cmark & \cmark & \cmark \\

\bottomrule
\end{tabular}}

\label{tab:gui_benchmark_comparison}
\end{table*}

\subsection{TVWorld-N}
\label{subsec:offline-simulated-device-dataset}

Based on the TV navigation graphs collected from Google TV, we construct \textbf{TVWorld-N}, an offline interactive TV navigation environment designed for comprehensive evaluation of agents' TV navigation capability. A comparison with mainstream GUI benchmarks is reported in Table~\ref{tab:gui_benchmark_comparison}.

\paragraph{Offline Interactive Environment.}
TVWorld-N provides a reproducible offline interactive TV navigation environment by leveraging our high-fidelity data collection pipeline, which establishes a one-to-one correspondence between real TV device states and graph nodes, as well as between device-level state transitions and graph edges. Through this construction, a complex real-world TV interaction environment that typically requires physical hardware, heavy software stacks, and non-trivial deployment is faithfully and almost entirely preserved in a lightweight, static graph representation. Consequently, evaluation can be conducted using only static assets, without running or maintaining interactive systems such as operating systems, virtual machines, mobile emulators, or browser automation frameworks. The static nature of the environment further guarantees reproducibility and enables millisecond-level interaction during evaluation.

\paragraph{Task Construction.}\label{sec:tvworld-n}
We formulate the topology-aware navigation task as a goal-directed remote-control navigation episode. Given an instruction that specifies a target page, the agent starts from an initial page and iteratively generates actions to interact with the dynamic environment, exploring the UI and moving the focus until the target node is reached. For each task, we randomly sample two distinct nodes from the graph as the start page and the goal page. For the goal specification, we consider two complementary formats: a textual goal defined by the name of the target node, and a visual goal represented by a screenshot of the target UI state. Accordingly, we construct both text-based instructions, such as ``I want to go to Privacy–Microphone page,” and vision-based instructions, such as ``<image>Navigate to the page shown in the image.” For each graph, we sample $50$ unique start–goal node pairs, resulting in a total of $500$ TV navigation tasks across all graphs.

\subsection{TVWorld-G}
\label{subsec:Focus-Aware}

Based on TVWorld, we construct a focus-aware grounding dataset termed \textbf{TVWorld-G}. For each node in the Google TV navigation graphs, we parse the corresponding view-tree and extract the bounding box of the currently focused element, represented as $(x_1, y_1, x_2, y_2)$ in screen coordinates, where $(x_1, y_1)$ and $(x_2, y_2)$ denote the top-left and bottom-right corners, respectively. All extracted annotations are manually verified and corrected when necessary to ensure quality. Finally, TVWorld-G contains $187$ samples for grounding evaluation.

\section{Topology-Aware Training}
\label{Sec:Method}

\begin{figure*}[h]
    \centering
    \includegraphics[width=0.85\textwidth]{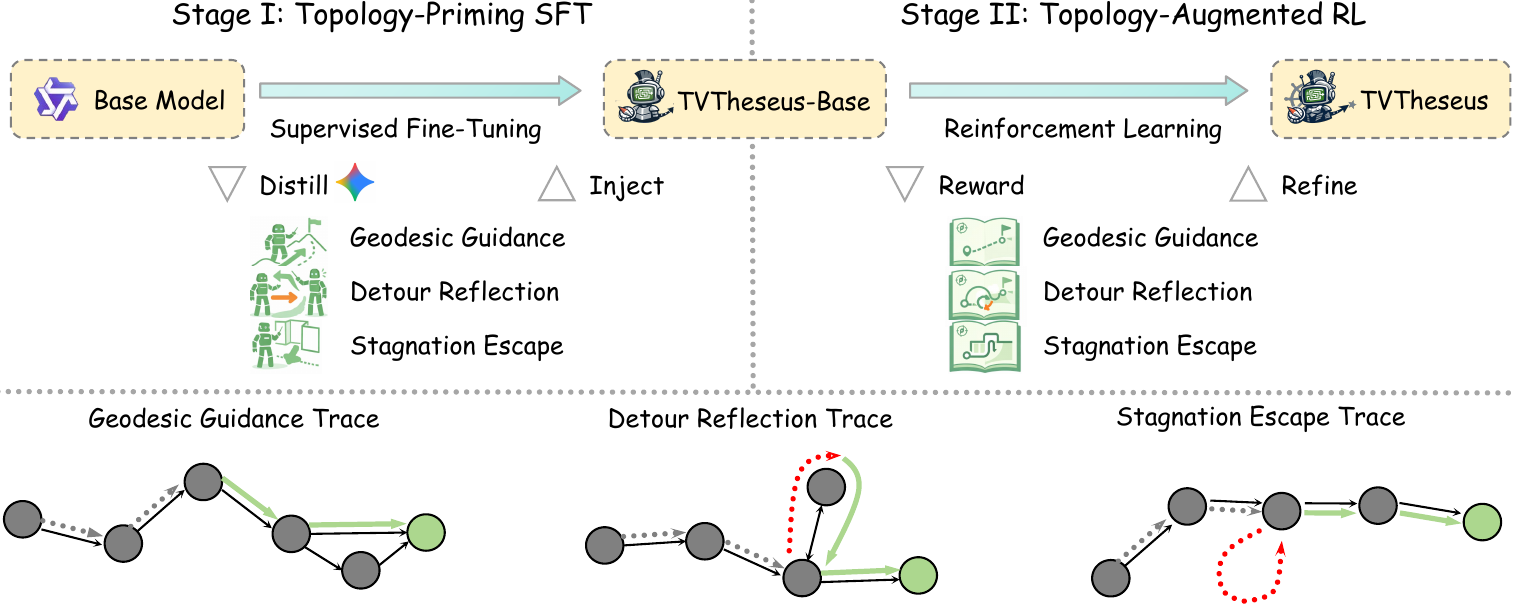}
    \caption{Overview of topology-aware training for TVTheseus. Stage I uses topology-priming SFT by distilling topology-aware behaviors and injecting them into the base model using three trace types: geodesic guidance, detour reflection, and stagnation escape. Stage II then applies topology-augmented RL with trace-specific rewards that promote goal-directed progress while discouraging detours and stagnation; example traces appear at the bottom.}
    \label{fig:trainillustfig}
\end{figure*}

Dependable TV navigation requires TV-use agents to reason over focus-based UI transitions in a goal-directed manner, while remaining robust to navigation errors such as detours and stalled states. We collectively refer to this interaction-level competence as \emph{topology awareness}. To embed this latent capability into TV-use agents, we introduce a two-stage training approach that first injects topology-aware inductive biases via topology-priming supervised fine-tuning (Sec.~\ref{sec:stage1}), and then progressively consolidates them through topology-augmented reinforcement learning (Sec.~\ref{sec:Topology Augmented RL}), as illustrated in Fig.~\ref{fig:trainillustfig}. Through this training paradigm, we obtain \textbf{TVTheseus}, a foundation model specialized for robust and generalizable TV control.

\subsection{Stage I: Topology-Priming SFT}
\label{sec:stage1}

In our early experiments, we observe that existing open-source models often fail to exhibit \emph{topology awareness}, leading to brittle behavior in focus-based TV navigation. To address this, we leverage natural-language reasoning~\cite{lu2025guiodyssey, wang2025opencua} as a mechanism for shaping the agent’s internal understanding of UI dynamics. Concretely, at each time step $t$, we constrain the model to produce a paired output $(z_t^{\star}, a_t^{\star})$, where $a_t^{\star}$ denotes the reference key action and $z_t^{\star}$ provides structured reasoning for that action. Building on this formulation, we introduce a topology-aware priming framework that instills such reasoning through three types of step-level rationales: \textbf{Geodesic Guidance}, \textbf{Detour Reflection}, and \textbf{Stagnation Escape}. Below, we describe the construction of these traces and the synthesis of the corresponding rationales $z_t^{\star}$ for a given start–goal pair $(u, g)$:

\paragraph{Geodesic Guidance Traces.}
\label{sec:geodesic}
We follow a clean geodesic route on the navigation graph from a start node to a goal node by computing the shortest path $\mathbf{p}^{\star}$. 
For each step $t$ in $\mathbf{p}^{\star}$, we endow $z_t^{\star}$ with (i) a description of the current UI state and focused element and (ii) an explanation of a locally plausible move that makes progress toward the goal. Instead of encouraging explicit topology memorization, these traces emphasize learning state transitions under key presses and maintaining an explicit notion of goal-directed progress.

\paragraph{Detour Reflection Traces.}
\label{sec:detour}

Real-world TV navigation often deviates from the shortest path, as an incorrect key press may move the agent farther from the goal, resulting in a \emph{topological detour}.
For a node $u_t$, we define a detour action $a_{\text{far}}$ as any action that increases the shortest-path distance to the goal, i.e., $d_{\text{sp}}(T(u_t, a_{\text{far}}), g) > d_{\text{sp}}(u_t, g)$. Starting from the shortest path, we deliberately insert such a detour and then return to the original node before continuing: 
$u_t \xrightarrow{a_{\text{far}}} u_{\text{far}} \xrightarrow{a_{\text{back}}} u_t \xrightarrow{a_t^{\star}} u_{t+1}$.
For action $a_t^{\star}$, we design $z_t^{\star}$ to reflect on the detour and justify a corrected move, explicitly discouraging repeating $a_{\text{far}}$ and favoring an action consistent with goal-directed progress.

\paragraph{Stagnation Escape Traces.}
\label{sec:stagnation}

We observe another common failure mode in TV navigation, where certain key presses do not trigger any state change. In such cases, the agent may repeatedly issue the same invalid key and become trapped in a local loop. To capture this behavior, we insert an invalid action ${a_{\text{inv}}}$ into an otherwise valid navigation segment:
 $u_t \xrightarrow{a_{\text{inv}}} u_t \xrightarrow{a_t^{\star}} u_{t+1}
$.
At this point, $z_t^{\star}$ is designed to recognize that the UI remains unchanged and to reason about the need to abandon the ineffective action, favoring an alternative key that leads to an actual transition. 

We use \texttt{Gemini 3 Pro Preview}~\cite{gemini3pro} to synthesize the three types of rationales $z_t^{\star}$; examples of the synthesized data are provided in Appendix~\ref{sec:Training Data Example}. Through supervised fine-tuning on these structured rationales, the base LVLM acquires foundational topology-aware capabilities for focus-based TV navigation. We refer to the resulting model after this stage as \textbf{TVTheseus-Base}.

\subsection{Stage II: Topology-Augmented RL}
\label{sec:Topology Augmented RL}

After topology-priming SFT equips the base LVLM with a strong initial policy, we introduce a second stage, Topology-Augmented Reinforcement Learning, to further consolidate topology-aware behaviors through interaction-driven optimization.

\subsubsection{Reinforcement Learning Formulation}

As introduced in Sec.~\ref{sec:tvworld}, TVWorld provides a fully offline, replayable interactive environment with millisecond-level response, enabling stable and efficient reinforcement learning without physical devices or online deployment. Within this environment, we construct interaction episodes on the training graphs following the same three trace patterns introduced in Stage I (Sec.~\ref{sec:stage1}), allowing the agent to explore alternative behaviors and receive feedback from state transitions.
We adopt GRPO~\citep{shao2024deepseekmath} as the optimization algorithm and initialize the policy from TVTheseus-Base. 
Details of GRPO are provided in Appendix~\ref{app:grpoapp}.

\subsubsection{Topology-Aware Reward Design}
\label{sec:Topology aware rewards}

In remote-control TV navigation, each valid key press induces a transition along an edge of the TV navigation graph, potentially moving the interface closer to, unchanged from, or farther away from the goal node. We exploit this structural property to design \emph{topology-aware rewards}. These rewards guide the agent’s behavior by examining the change in graph distance to the goal between the current node $u_t$ and the node $u'$ reached after executing a model-suggested action, that is, $d(u_t,g)-d(u',g)$.

\paragraph{Trace-Specific Reward Design.}
Our shaping reward assigns higher values to actions that make goal-directed progress (reducing $d(\cdot,g)$), a small positive value to distance-preserving moves, and lower values to actions that move away from the goal. 
We further incorporate trace-dependent penalties to correct common failure modes: \emph{Detour Reflection} discourages returning to the previously identified detour branch, while \emph{Stagnation Escape} discourages repeating the invalid action. Complete reward definitions are provided in Appendix~\ref{app:trace-reward}.

\paragraph{Distance Definition.}
We instantiate the distance metric as the \emph{hitting time} $d(u,g)=h_g(u)$, which measures the expected number of steps for a random walk starting from node $u$ to first reach the target node $g$. Let $A$ denote the adjacency matrix of the navigation graph, where $A_{uv}$ counts feasible actions from $u$ to $v$, $D=\mathrm{diag}(\sum_v A_{uv})$, and $P=D^{-1}A$. With $h_g(g)=0$, the hitting time satisfies
$
h_g(u)=1+\sum_v P_{uv}\,h_g(v).
$
Detailed distance definitions and alternatives are provided in Appendix~\ref{app:distances}.

We combine the topology-aware reward with a format-validity reward into a single scalar objective: $R=\beta_{\text{topo}}R_{\text{topo}}+\beta_{\text{form}}R_{\text{form}}.$ Together, these rewards provide dense and structured feedback that reinforces topology-aware behaviors without relying on fixed reference actions or auxiliary verifiers~\cite{bonatti2024windows, chen2025gui,devidze2021explicable}. By aligning reinforcement learning objectives with the topology-aware reasoning patterns introduced during Stage I, this design further unlocks the latent topology awareness of the LVLM, yielding policies that are more robust and better reflect real-world remote-control navigation.

\section{Experiments}
\subsection{Experimental Setup}

\paragraph{Training Settings.}
We adopt Qwen3-VL-8B-Instruct~\cite{Qwen3-VL} as the base model. At each time step $t$, the input is
$\mathbf{x}_t = (S_t, S_{t-1:t-\delta_S}, a_{t-1:t-\delta_a}, I)$,
including the current screenshot $S_t$, up to $4$ historical screenshots ($\delta_S=4$), the full action history ($\delta_a=t$), and the instruction $I$. All experiments are conducted on $8\times$ NVIDIA A100 GPUs. Further details on the training data and the two-stage training setup are provided in Appendix~\ref{app:Training data description} and Appendix~\ref{app:Training Setup Details}, respectively.

\paragraph{Evaluation.}
We evaluate our TVTheseus on two tasks: Topology-aware Navigation and Focus-aware Grounding.

\textbf{Topology-aware Navigation.}
Evaluation is conducted on TVWorld-N, which consists of $5$ navigation graphs, each containing $100$ tasks with both text-based and vision-based instructions (Sec.~\ref{sec:tvworld-n}), totaling $500$ tasks. We compare TVTheseus with (i) closed-source models (GPT-5 mini~\cite{gpt5}, Gemini 3 Flash~\cite{gemini3pro}, Claude Haiku 4.5~\cite{claudehaiku}), (ii) general-purpose open-source LVLMs (Qwen3-VL-8B-Instruct, Qwen3-VL-32B-Instruct~\cite{Qwen3-VL}), and (iii) pointer-based UI control models (UI-Tars-1.5-7B~\cite{ui-tars-15-seed}, OpenCUA-7B~\cite{wang2025opencua}, GUI-Owl-7B, GUI-Owl-32B~\cite{ye2025mobile}). All models are evaluated with a maximum horizon of 50 steps, image resolution $1024\times576$, up to 4 historical screenshots, and the full action history. We report \textbf{Success Rate (SR)}, defined as finishing on the target page.

\textbf{Focus-aware Grounding.}
We evaluate focus localization on TVWorld-G, which contains $187$ samples, comparing TVTheseus with general-purpose LVLMs (Qwen2.5-VL-7B-Instruct, Qwen3-VL-8B-Instruct, Qwen3-VL-8B-Thinking~\cite{Qwen3-VL}) and pointer-based grounding models (InfiGUI-R1-3B~\cite{liu2025infigui}, GUI-R1-3B, GUI-R1-7B~\cite{luo2025gui}, GUI-Owl-7B, GUI-Owl-32B~\cite{ye2025mobile}). All models use an input resolution of $1024\times576$. Performance is measured by $\textbf{\text{Acc@0.5}}$ (IoU $\ge 0.5$).

\subsection{Main Results}

\paragraph{Interactive TV Navigation Evaluation.}
Table~\ref{tab:main_nav} reports results on TVWorld-N. TVTheseus achieves the best overall performance, outperforming all baselines, including the strongest closed-source model, Gemini 3 Flash. On previously unseen TV platforms, TVTheseus markedly improves over its base model, Qwen3-VL-8B-Instruct, with success rate increasing from $20.0$ to $68.3$. Appendix~\ref{appx: case_study} provides a qualitative case study illustrating the behavioral differences before and after training. This result demonstrates strong out-of-domain generalization enabled by our two-stage Topology-Aware Training.
In comparison, the strongest general-purpose open-source model (i.e., Qwen3-VL-32B-Instruct) reaches a success rate of $39.0$, substantially lagging behind closed-source models such as Gemini 3 Pro and GPT-5 mini. Models trained for point-and-click (PnC) interaction (e.g., GUI-Owl and OpenCUA) degrade substantially in the remote-control (RC) TV setting, highlighting the mismatch between pointer-based assumptions and focus-based TV navigation.
Performance also varies consistently across scenarios: most models perform better on \textit{Apps} and \textit{Channels}, while \textit{Config} remains the most challenging. This pattern suggests that TVWorld-N is a discriminative benchmark that effectively distinguishes genuine topology-aware navigation from superficial interaction heuristics.
\begin{table*}[h]
\tiny
\renewcommand{\arraystretch}{0.5}
\centering
\caption{Comprehensive evaluation on TVWorld-N across five out-of-domain scenarios. Boldface indicates the best performance. For \textbf{TVTheseus}, we report the mean over three independent runs. Detailed results for each individual run are provided in Appendix Table~\ref{tab:all_result}.}
\scalebox{1.2}{
\begin{tabular}{l c ccccc c}
\toprule
Model & Instr Type & Config & Display & Audio & Apps & Channels & Overall \\
\midrule
\rowcolor{gray!15}
\multicolumn{8}{l}{} \\
\multirow{2}{*}{GPT-5 mini} 
  & text-based & $\textbf{48.0}$ & 54.0 & 58.0 & 48.0 & 68.0 & \multirow{2}{*}{60.2} \\
  & vision-based & 60.0 & 54.0 & 52.0 & 78.0 & 82.0 &  \\
\cmidrule(lr){1-8}
\multirow{2}{*}{Gemini 3 Flash} 
  & text-based & 46.0 & $\textbf{66.0}$ & 58.0 & 62.0 & 74.0 & \multirow{2}{*}{66.4} \\
  & vision-based & 54.0 & $\textbf{68.0}$ & $\textbf{62.0}$ & 80.0 & $\textbf{94.0}$ &  \\
\cmidrule(lr){1-8}
\multirow{2}{*}{Claude Haiku 4.5}
  & text-based & 24.0 & 34.0 & 42.0 & 40.0 & 50.0 & \multirow{2}{*}{25.4} \\
  & vision-based &  2.0 & 10.0 & 12.0 & 20.0 & 20.0 &  \\
\midrule
\rowcolor{gray!15}
\multicolumn{8}{l}{\textit{General Open-source Model}} \\
\multirow{2}{*}{Qwen3-VL-8B-Instruct}
  & text-based &  8.0 & 24.0 & 10.0 & 40.0 & 16.0 & \multirow{2}{*}{20.0} \\
  & vision-based & 14.0 & 10.0 & 10.0 & 32.0 & 36.0 &  \\
\cmidrule(lr){1-8}
\multirow{2}{*}{Qwen3-VL-32B-Instruct}
  & text-based & 26.0 & 30.0 & 44.0 & 34.0 & 44.0 & \multirow{2}{*}{39.0} \\
  & vision-based & 22.0 & 42.0 & 36.0 & 60.0 & 52.0 &  \\
\midrule
\rowcolor{gray!15}\multicolumn{8}{l}{\textit{PnC-specific model}} \\
\multirow{2}{*}{UI-Tars-1.5-7B}
  & text-based &  0.0 &  4.0 &  2.0 &  6.0 &  0.0 & \multirow{2}{*}{1.6} \\
  & vision-based &  0.0 &  0.0 &  0.0 &  4.0 &  0.0 &  \\
\cmidrule(lr){1-8}
\multirow{2}{*}{OpenCUA-7B}
  & text-based &  0.0 &  2.0 &  8.0 &  4.0 & 18.0 & \multirow{2}{*}{5.0} \\
  & vision-based &  0.0 &  2.0 &  6.0 &  6.0 &  4.0 &  \\
\cmidrule(lr){1-8}
\multirow{2}{*}{GUI-Owl-7B}
  & text-based &  0.0 &  2.0 & 10.0 &  2.0 & 16.0 & \multirow{2}{*}{5.4} \\
  & vision-based &  2.0 &  0.0 &  2.0 & 10.0 & 10.0 &  \\
\cmidrule(lr){1-8}
\multirow{2}{*}{GUI-Owl-32B}
  & text-based &  2.0 & 18.0 & 22.0 & 20.0 & 26.0 & \multirow{2}{*}{15.4} \\
  & vision-based &  4.0 & 10.0 &  6.0 & 20.0 & 26.0 &  \\
\midrule

\rowcolor{gray!15}\multicolumn{8}{l}{\textit{RC-specific model}} \\
  \multirow{2}{*}{\textbf{TVTheseus (Ours)}}  
  & text-based & $\text{41.3}^{+6.7}_{-5.3}$ & $\text{62.7}^{+9.3}_{-10.7}$ & $\textbf{86.0}^{+2.0}_{-4.0}$ & $\textbf{70.7}^{+5.3}_{-2.7}$ & $\textbf{76.0}^{+8.0}_{-4.0}$ & \multirow{2}{*}{$\textbf{68.3}^{+3.1}_{-2.7}$} \\
  & vision-based & $\textbf{62.7}^{+3.3}_{-4.7}$ & $\text{60.7}^{+3.3}_{-6.7}$ & $\text{58.0}^{+6.0}_{-10.0}$ & $\textbf{82.7}^{+1.3}_{-0.7}$ & $\text{82.7}^{+3.3}_{-4.7}$ &  \\
\bottomrule
\end{tabular}
}
\label{tab:main_nav}
\end{table*}

\paragraph{Focus-Aware Grounding Evaluation.}
Table~\ref{tab:main_gr} reports focus-aware grounding performance on TVWorld-G. Although TVTheseus is not trained with any grounding-specific supervision, it outperforms its base model (i.e., Qwen3-VL-8B-Instruct), by $3.7$ points, achieving the best overall Acc@0.5 of $81.8$. This result indicates that strong topology awareness acquired in TV environments transfers to improved focus localization.
We also observe that Qwen3-VL-8B-Thinking performs $8.6$ points worse than Qwen3-VL-8B-Instruct, suggesting that explicit multi-step reasoning may not be necessary for this task. Consistent with navigation results, PnC-specific models underperform general-purpose LVLMs, further reflecting the fundamental mismatch between pointer-based and remote-control interaction paradigms, which impose distinct capability requirements on LVLM agents.

\begin{table}[h]
\centering
\caption{Focus-awareness performance on TVWorld-G.}

\renewcommand{\arraystretch}{0.9}
\tiny
\scalebox{1.5}{
\begin{tabular}{l  c}
\toprule
Model & Acc@0.5  \\
\midrule
\rowcolor{gray!15}
\multicolumn{2}{l}{\textit{General Open-source Model}} \\
Qwen2.5-VL-7B-Instruct   & 66.3 \\
Qwen3-VL-8B-Instruct  &  78.1 \\
Qwen3-VL-8B-Thinking & 69.5 \\
\midrule
\rowcolor{gray!15}\multicolumn{2}{l}{\textit{PnC-specific model}} \\
InfiGUI-R1-3B  & 56.7 \\
GUI-R1-3B  & 49.2 \\
GUI-R1-7B  & 65.2 \\
GUI-Owl-7B & 39.5 \\
GUI-Owl-32B  & 54.0 \\
\midrule
\rowcolor{gray!15}\multicolumn{2}{l}{\textit{RC-specific model}} \\
\textbf{TVTheseus (Ours)} & \textbf{81.8} \\

\bottomrule
\end{tabular}}
\label{tab:main_gr}
\end{table}

\subsection{Ablation Study}

\paragraph{Effect of the Topology-Aware Training Strategy.}
Table~\ref{tab:ablation_stage} shows that the two training stages play complementary roles. Stage I (SFT) establishes core topology-aware behaviors and strong focus awareness, while Stage II (RL) further improves long-horizon planning and recovery. Although Stage II introduces a mild trade-off in focus grounding, the impact is limited, and the model gains substantially stronger topology-aware navigation capability.

\begin{table}[h]
\centering
\caption{Impact of two-stage topology-aware training.}

\renewcommand{\arraystretch}{0.6}
\tiny
\scalebox{1.4}{
\begin{tabular}{l c c}
\toprule
Training Strategy & TVWorld-N & TVWorld-G  \\
\midrule
-- & 20.0 & 78.1 \\
\quad+Stage I & 48.0 & 83.4 \\
\quad+Stage I \& Stage II & 68.3 & 81.8\\
\bottomrule
\end{tabular}}
\label{tab:ablation_stage}
\end{table}

\paragraph{Additional Experiments.}

We report additional experiments in Appendix~\ref{appx: more_exp}. These include ablations on rationale types, distance definitions, and reward formulations, as well as analyses of how image resolution and the number of historical screenshots affect model performance.

\section{Conclusion}
In this work, we present TVWorld, a comprehensive and static interactive resource that fills a critical gap in remote-control–based TV agent development. By providing a unified set of benchmarks, an effective training framework, and a specialized TV foundation model, we establish essential building blocks for studying TV-use agents under remote-control interaction paradigms. We hope this work will spur further study of GUI agents in remote-control settings and catalyze broader research on this interaction paradigm.

\section*{Limitations}
This work focuses on TV-use agents at a practical model scale that is representative of current deployable systems, rather than performing an exhaustive scaling study on substantially larger pretrained models. This design choice enables controlled and systematic analysis of topology-aware training and evaluation under realistic computational budgets. While we do not explore scaling effects, the core phenomena and conclusions regarding topology-aware behavior are not tied to model size and are therefore expected to generalize.
In addition, as part of a responsible data collection and release process, we mask a small number of sensitive system-level entry points during graph construction. As a result, the constructed navigation graphs differ slightly from real-world TV deployments. This masking is a deliberate measure to ensure data quality and safe release, while preserving the core interaction structure, navigation topology, and focus-based control dynamics.

\bibliography{custom}

\clearpage

\appendix

\section{Related Work}
\label{sec:Related work}
\paragraph{Graph-based methods.} Graphs provide a convenient abstraction for different domains, including materials science~\cite{butler2018machine,you2018graph}, engineering~\cite{darvariu2021goal,yang2023learning}, and networking security~\cite{nyberg2023training,xu2022moving}. Specifically, robotics frames motion planning as search on configuration graphs~\cite{hossain2024toponav,shah2021ving}. Knowledge-graph reasoning casts question answering as multi-hop traversal~\cite{xiong2017deeppath,das2017go}. For GUI agents, interfaces are often represented as DOM or state graphs, where edges correspond to actionable elements~\cite{jia2019dom,gur2018learning,adamo2018reinforcement,pan2020reinforcement,zhang2025progrm,xu2024crab}; In the TV auto testing domain, prior work uses crawlers to construct UI graphs to generate test sequences~\cite{firat2022model,ahmed2020automated,bures2020testing}. There has been almost no systematic exploration of training LVLMs with graph-based reinforcement learning in remote-control scenarios.

\paragraph{LVLM for GUI Control.}
Work on GUI agents spans the web~\cite{zhang2025litewebagent,abuelsaad2024agent}, mobile device~\cite{li2024effects,papoudakis2025appvlm}, and desktop control~\cite{zhang2025ufo2,zhao2025cola}. One research direction enhances inputs with A11y trees~\cite{wu2024atlas}, Set-of-Marks~\cite{agashe2024agent}, or DOM~\cite{schiepanski2025beyond} to supply models with fine-grained UI details. Another approach employs control based solely on screenshots to directly determine action positions from pixels~\cite{hong2024cogagent,li2024ferret,shaw2023pixels,wang2024ponder,gou2024navigating,ye2025mobile,chen2024gui}. This method is versatile across tasks and devices, and particularly useful when structural inputs are absent or impractical~\cite{cheng2024seeclick,xie2024osworld}, making it a promising long-term path for transferability. Despite these advances, current LVLM agents still operate mainly in a point-and-click paradigm on cursor or touch interfaces, while remote control scenarios such as TVs remain largely underexplored.
\paragraph{Training Methods for GUI Control.}
GUI-control agents are commonly trained with supervised fine-tuning (SFT), often augmented with chain-of-thought, to improve action prediction~\cite{baechler2024screenai,ye2025mobile,lu2025steve,zhang2024android,zhang2025does}. Beyond SFT, which depends on large annotated datasets, recent work frames GUI control as a reinforcement-learning problem via reward design and policy optimization~\cite{lu2025ui,luo2025gui,liu2025infigui,zhou2025gui,lee2025reguide,lu2025arpo,li2025webthinker}, enabling greater sample efficiency and stronger generalization to novel tasks. Complementary efforts explore multi-agent RL and the integration of external tools~\cite{lu2025swirl,singh2025agentic,zeng2025reinforcing}.

\section{Action Set}
\label{app:details of action_set}
TVWorld supports $8$ discrete button actions. We further introduce a \texttt{FINISH} action to signal task completion, resulting in a total of $9$ actions in the action space. The complete action space and their functionalities are detailed in Table~\ref{tab:action-space}.

\begin{table*}[]
    \centering
    \caption{The functionality of different actions in TVWorld.}
    \label{tab:action-space}
    \begin{tabular}{c| c}
    \toprule
        \textbf{Action}  & \textbf{Functionality} \\
        \midrule
        \texttt{UP} &  move the focus upward\\
        \midrule
        \texttt{DOWN} &  move the focus downward  \\
        \midrule
        \texttt{LEFT} &  move the focus to the left or return to the parent directory menu \\
        \midrule
        \texttt{RIGHT} &  move the focus to the right or enter the highlighted item \\
        \midrule
        \texttt{OK} & confirm the current selection or enter a highlighted item \\
        \midrule
        \texttt{HOME} & return to the home screen\\
        \midrule
        \texttt{EXIT} & exit the current page or return to the parent directory menu\\
        \midrule
        \texttt{SETTING} & open the settings screen \\
        \midrule
        \texttt{FINISH} & indicate that the navigation task is completed \\
        \bottomrule
    \end{tabular}
\end{table*}

\section{Training Setup Details}
\label{app:Training Setup Details}

\paragraph{Stage I (SFT).} We randomly sample $500$ traces from the TCL TV navigation graph, yielding $8{,}490$ training instances (6,490 Geodesic Guidance, 1,000 Detour Reflection, and 1,000 Stagnation Escape). Training uses the official Qwen3-VL codebase~\cite{Qwen3-VL} with DeepSpeed ZeRO-1, a learning rate of $1\times10^{-6}$, global batch size $64$, weight decay $0$, and a maximum of $589{,}824$ vision tokens. Training runs for $5$ epochs (about $27$ GPU hours).

\paragraph{Stage II (RL).} We sample $1{,}000$ traces, resulting in $17{,}825$ training instances (11,825 Geodesic Guidance, 4,000 Detour Reflection, and 2,000 Stagnation Escape). Reward weights are set to $\beta_{\text{topo}}=0.95$ and $\beta_{\text{form}}=0.05$. Training is performed with the VeRL framework~\cite{sheng2025hybridflow} and vLLM~\cite{kwon2023efficient}, using a global batch size of $64$, rollout size $8$, and $600$ optimization steps (about $280$ GPU hours).

\section{Training Data Description}
\label{app:Training data description}

%

A path from \(u\) to \(g\) is denoted by \(\mathbf{p}=(u_0,a_0,u_1,\ldots,u_{L-1},a_{L-1},u_L)\) with \(u_0=u\), \(u_L=g\). Its length is \(\operatorname{len}(\mathbf{p})=L\), and \(\Pi(u\!\to\! g)\) denotes the set of all finite paths from \(u\) to \(g\). We denote the shortest path connected $u$ and $g$ as $\underset{\mathbf{p}\in \Pi(u\!\to\! g)}{\arg\min}\ \mathrm{len}(\mathbf{p})$.

Our training data graph $\mathcal{G}$ is collected from TCL TV. 
We select start--goal pairs $(u_0,g)$ and build a path,
\begin{equation}
\mathbf{p}^{\star}
=\big(u^{\star}_0=u_0, a^{\star}_0, \dots, u^{\star}_L=g\big).
\end{equation}
From each timestep $t$ on $\mathbf{p}^{\star}$, it can form a training sample
\begin{equation}
\begin{aligned}
\xi_t &= \Big(S_t,~H_t,~I\Big),\qquad S_t = S\!\big(u_t^{\star}\big),\\
H_t &= \big(a^{\star}_{t-\delta_{a}},\dots,a^{\star}_{t-1},\, S_{t-\delta_{S}},\dots,S_{t-1}\big),
\end{aligned}
\end{equation}
where $S_t$ is the current screenshot, $H_t$ concatenates the last $\delta_{a}$ actions and $\delta_{S}$ screenshots, and $I$ is an instruction specifying the final goal $g$.

\section{Trace-Specific Topology-Aware Reward Functions}
\label{app:trace-reward}
This Appendix section details the trace-specific topology-aware shaping rewards used in Stage II. For a state–action pair $(u_t, a)$ within a given trace, the reward is computed from the resulting node $u'$, reached after taking action $a$ at $u_t$, by comparing the graph-based distances $d(u', g)$ and $d(u_t, g)$. In this way, each reward component reinforces its corresponding topology-aware behavior through goal-directed progress.

For \textbf{Geodesic Guidance Traces}, we directly encourage topology-consistent progress by favoring actions that reduce the distance to the goal:
\begin{equation}
\label{eq:geo-reward}
{\small
R_{\text{geo}}(u_{t},a;g)=
\begin{cases}
1, & d(u',g) < d(u_{t},g),\\
0.2, & d(u',g) = d(u_{t},g),\\
0, & d(u',g) > d(u_{t},g),
\end{cases}
}
\end{equation}

For \textbf{Detour Reflection Traces}, at the revisited node $u_t$, we preserve the preference for moving closer to the goal while explicitly discouraging returning to the previously identified detour branch:
\begin{equation}
\label{eq:detour-reward}
{\small
R_{\text{det}}(u_{t},a;g)=
\begin{cases}
1, & d(u',g) < d(u_{t},g),\\
0.2, & d(u',g) = d(u_{t},g),\\
0.1, & d(u',g) > d(u_{t},g),\; a\neq a_{\text{far}},\\
0, & d(u',g) > d(u_{t},g),\; a= a_{\text{far}}.
\end{cases}
}
\end{equation}

For \textbf{Stagnation Escape Traces}, we focus on the second visit to $u_t$ following an invalid key press. Let $a_{\text{inv}}$ denote the stagnating action. The reward penalizes repeating $a_{\text{inv}}$ while continuing to shape behavior toward goal-directed progress:
\begin{equation}
\label{eq:stagnation-reward}
{\small
R_{\text{sta}}(u_{t},a;g)=
\begin{cases}
1, & d(u',g) < d(u_{t},g),\\
0.2, & d(u',g) = d(u_{t},g),\; a\neq a_{\text{inv}},\\
0, & d(u',g) = d(u_{t},g),\; a= a_{\text{inv}},\\
0.1, & d(u',g) > d(u_{t},g).
\end{cases}
}
\end{equation}
\section{Distance Families for Topological Shaping}
\label{app:distances}
This Appendix section formalizes several commonly used graph-distance functions \(d(\cdot,g)\) that capture the notion of topological proximity and can be used to construct topology-aware rewards.

Recall that Eqs.~(\ref{eq:geo-reward}-\ref{eq:stagnation-reward}) reward an action precisely through how it changes the distance to the goal node \(g\).
Therefore, we introduce a few graph-based distance families that can serve as \(d(\cdot,g)\), using a unified notation throughout.
Let the TV navigation graph be a labeled directed multigraph \(\mathcal{G}=(\mathcal{V},\mathcal{E},\lambda)\) with a transition map \(T:\mathcal{V}\times\mathcal{A}\to\mathcal{V}\), and let \(n\triangleq|\mathcal{V}|\).

Define the adjacency matrix \(A\in\mathbb{R}^{n\times n}\) by
\[
A_{uv}\triangleq \bigl|\{a:\,(u,a,v)\in\mathcal{E}\}\bigr|,
\]
so \(A_{uv}\) counts the number of labeled edges from \(u\) to \(v\).
Let \(A_{\mathrm{rev}}\triangleq A^\top\), i.e., \((A_{\mathrm{rev}})_{uv}=A_{vu}\).
Let \(e_g\in\mathbb{R}^{n}\) denote the standard basis vector with a \(1\) at the coordinate corresponding to node \(g\). In all experiments, the TV state-transition graph is strongly connected. Below are four common distance definitions. 
\subsection{Shortest-Path Distance}
A natural choice is the directed shortest-path distance
\[
d_{\mathrm{sp}}(u,g)\triangleq \underset{\mathbf{p}\in \Pi(u\!\to\! g)}{\min}\ \mathrm{len}(\mathbf{p}),
\]
which measures the minimum number of actions required to reach the goal node \(g\) from state \(u\) along directed transitions.

\subsection{Hitting Time}
\label{app:hitting-time}

\textbf{Definition.}
Let
\[
D \;=\; \mathrm{diag}\!\Big(\sum_{v} A_{uv}\Big),\qquad
P \;=\; D^{-1}A
\]
be the forward row-stochastic random-walk matrix that chooses uniformly among feasible labeled edges. Make $g$ absorbing by replacing row $g$ of $P$ with $e_g^\top$ (so $P_{gg}=1$, $P_{gv}=0$ for $v\neq g$).
Let $\bar g \triangleq \mathcal{V}\setminus\{g\}$ and write the block
$Q \triangleq P_{\bar g\,\bar g}$ as the $(n\!-\!1)\times(n\!-\!1)$ submatrix obtained by deleting the row and column of $g$ in $P$. The hitting-time vector $h_g\in\mathbb{R}^{n}$ is the solution to the Dirichlet problem with $h_g(g)=0$ and
\[
h_g(u) \;=\; 1 + \sum_{v} P_{uv}\, h_g(v)\ \ \text{for } u\neq g.
\]
Equivalently,
\[
\big(I - Q\big)\, h_g(\bar g) \;=\; \mathbf{1},\qquad h_g(g)=0.
\]
We set
\[
d_{\mathrm{hit}}(u,g) \;\triangleq\; h_g(u).
\]
\paragraph{Interpretation.}
$d_{\mathrm{hit}}(u,g)$ is the expected number of remote steps required by an uninformed random policy to reach $g$ from $u$.
It therefore reflects exploration difficulty: narrow funnels, dead ends, and high-branching detours inflate $d_{\mathrm{hit}}$ even when $d_{\mathrm{sp}}$ is small~\citep{blum2020foundations}.
\subsection{Soft Shortest-Walk}
\paragraph{Definition.}
Let $A\in\mathbb{R}^{n\times n}$ be the adjacency matrix. For a temperature $\beta>0$, define the discounted adjacency
\begin{equation}
W \triangleq e^{-\beta}A,\qquad
Z \triangleq (I-W)^{-1} = \sum_{k=0}^{\infty} W^{k},
\end{equation}
where we assume $\rho(W)<1$ so the Neumann series converges. We define the soft shortest-walk distance
\begin{equation}
d_{\text{soft}}(u,g) \triangleq -\frac{1}{\beta}\log Z_{ug}.
\label{eq:softsp}
\end{equation}
Since $(W^k)_{ug}=e^{-\beta k}(A^k)_{ug}$, we have
\begin{equation}
Z_{ug}=\sum_{k\ge 0} e^{-\beta k}\,(A^{k})_{ug}
      =\sum_{\pi:u\to g}\exp(-\beta|\pi|),
\end{equation}
where $\pi$ ranges over all (action-labeled) walks from $u$ to $g$ and $|\pi|$ is its length. Therefore, $d_{\text{soft}}(u,g)$ can be viewed as a log-sum-exp relaxation of the shortest-walk length over all walks.

\paragraph{Interpretation.}
Let $m\triangleq \min\{k\ge 1:(A^{k})_{ug}>0\}$ be the shortest-walk length from $u$ to $g$. Then,
{\small
\begin{equation}
d_{\text{soft}}(u,g)
= m-\frac{1}{\beta}\log\!\Big(\sum_{t\ge 0} (A^{m+t})_{ug}\,e^{-\beta t}\Big)
\le m.
\end{equation}
}
Walks that are $k$ steps longer receive at most a relative weight $e^{-\beta k}$, so sufficiently longer walks are exponentially suppressed. In particular, if there are $N_m=(A^{m})_{ug}$ shortest walks and longer walks contribute little, then
$d_{\text{soft}}(u,g)\approx m-\frac{1}{\beta}\log N_m$.
Moreover, increasing any entry of $A$ (e.g., adding edges or increasing counts) can only increase $Z_{ug}$, and thus can only decrease $d_{\text{soft}}(u,g)$. Finally, since $(A^k)_{ug}>0$ iff there exists a directed walk of length $k$ from $u$ to $g$, $m$ equals the directed shortest-path length from $u$ to $g$. Hence $d_{\text{soft}}$ can be interpreted as a soft version of the shortest path: as $\beta\to\infty$, $d_{\text{soft}}(u,g)\to m$.

\subsection{Personalized PageRank}

\textbf{Definition.}
Construct the forward row-stochastic random-walk matrix $P=D^{-1}A$ from the adjacency $A$. For each seed $u$, the personalized PageRank (PPR) vector $p_u\in\mathbb{R}^n$ solves
\[
p_u \;=\; \alpha\, e_u \;+\; (1-\alpha)\, P^{\!\top} p_u
\]
where $\alpha\in(0,1)$ is the restart probability.
We then define the forward PPR distance to target $g$ by
\[
d_{\mathrm{ppr}}(u,g) \;\triangleq\; 1 - p_u(g).
\]
\paragraph{Interpretation.}
$p_u$ denotes the stationary visit distribution of a random walk on the forward graph that, at each step, returns to the current-state seed $u$ with probability $\alpha$. Consequently, $p_u$ can be interpreted as an exponentially discounted combination of the $t$-step walk distributions originating from $u$, where the contribution of longer walks decays exponentially. In this way, $p_u(g)$ reflects the long-run visitation frequency of node $g$ under an uninformed exploration process rooted at $u$, making it a measure of proximity~\citep{tong2006fast,andersen2006local}. We then define the distance
$d_{\mathrm{ppr}}(u,g)=1-p_u(g)$, so that nodes that are visited more often are regarded as closer.
\section{Group-Relative Policy Optimization}
\label{app:grpoapp}

This appendix specifies the Group-Relative Policy Optimization (GRPO) objective used in the second stage of Topology-Aware Training (Sec.~\ref{sec:Topology Augmented RL}), where the agent is rewarded by topology-aware rewards derived from the TV navigation graph (Sec.~\ref{sec:Topology aware rewards}).

\paragraph{Structured generation and executable interface.}
For each training example, the agent takes $\xi_t=(S_t,H_t,I)$ as input (defined in Appendix~\ref{app:Training data description}) and generates a response that includes both a rationale and a single executable remote-control key:
\begin{gather}
{\small r_t=~z_t~~\texttt{<answer>}~a_t~\texttt{</answer>}\notag} \\
    {\small r_t\sim \pi_\theta(r_t\mid S_t,H_t,I)}.
\end{gather}
The action token $a_t$ is then executed in the environment. Concretely, a response is considered well-formed if (i) the tags are balanced, (ii) there is exactly one \texttt{<answer>} span. This makes output validity a learnable preference signal during training, without hard-coding a constrained decoder at test time.

\paragraph{Sampling a group and computing rewards.}
Fix the context $(S_t,H_t,I)$ at node $u_t^{\star}$. GRPO generates a {group} of $K$ candidate responses $\{r^{(k)}\}_{k=1}^{K}\sim\pi_\theta$, extracts their actions $\{a^{(k)}\}$, and applies one environment step transition to get $u'^{(k)}$. Each candidate is assigned two reward components:
\begin{equation}
R^{(k)}_{\text{topo}}=R_{\text{topo}}(u_t^{\star},a^{(k)};g),
\qquad
R^{(k)}_{\text{form}}\in\{0,1\},
\end{equation}
where $R_{\text{topo}}$ is the topology-aware shaping reward defined in Appendix Sec.~\ref{app:trace-reward} (with trace-specific instantiations such as $R_{\text{geo}},R_{\text{det}},R_{\text{sta}}$), and
$R^{(k)}_{\text{form}}=1$ iff $r^{(k)}$ is well-formed (balanced tags, exactly one \texttt{<answer>}).

We combine them into a single scalar score,
\begin{equation}
R^{(k)}=\beta_{\text{topo}}R^{(k)}_{\text{topo}}+\beta_{\text{form}}R^{(k)}_{\text{form}},
\end{equation}
so that the policy is simultaneously encouraged to (i) take keys that make measurable progress on the TV graph and (ii) emit reliably executable outputs.

\paragraph{Group-relative advantages.}
Unlike value-based methods, GRPO normalizes scores {within the sampled group} for the same context:
\begin{equation}
A^{(k)}=\frac{R^{(k)}-\mu}{\sigma},
\end{equation}
where $\mu$ and $\sigma$ are the mean and standard deviation of $\{R^{(k)}\}_{k=1}^K$. This turns raw rewards into a scale-free, variance-reduced advantage: candidates are compared only against their peers under the same $(S_t,H_t,I)$, which is precisely what we need when supervision comes from graph transitions rather than a single reference action.

\paragraph{GRPO objective loss.}
Let the $k$-th response be tokenized as $\{r^{(k)}_\ell\}_{\ell=1}^{|r^{(k)}|}$. Define the per-token importance ratio
\[
v^{(k)}_\ell=
\frac{\pi_\theta\!\left(r^{(k)}_\ell\mid S_t,H_t,I,r^{(k)}_{<\ell}\right)}%
{\pi_{\theta_{\text{old}}}\!\left(r^{(k)}_\ell\mid S_t,H_t,I,r^{(k)}_{<\ell}\right)}.
\]
GRPO optimizes a clipped surrogate and regularizes the policy toward a reference model $\pi_{\text{ref}}$ to prevent uncontrolled drift:
\begin{equation}
\bar v^{(k)}_\ell \triangleq
\mathrm{clip}\!\big(v^{(k)}_\ell,\,1-\epsilon,\,1+\epsilon\big).
\end{equation}

\begin{equation}
s^{(k)}_\ell \triangleq
\min\!\Big(v^{(k)}_\ell A^{(k)},\,\bar v^{(k)}_\ell A^{(k)}\Big).
\end{equation}
\begin{equation}
\begin{aligned}
\mathcal{L}_{\textsc{grpo}}
&=\mathbb{E}_{\substack{\xi\sim\mathcal{D}\\ r^{(k)}\sim\pi_\theta(\cdot\mid\xi)}}\frac{1}{K}\sum_{k}\frac{1}{|r^{(k)}|}\sum_{\ell}\{
[ s^{(k)}_\ell]\\
&\quad-\lambda_{\text{KL}}\,
\operatorname{KL}\!\left(\pi_\theta\,\|\,\pi_{\text{ref}}\right)\}.
\end{aligned}
\end{equation}

The $A^{(k)}$ is computed once for each sampled response and then applied to all of its tokens via the importance ratio. As a result, the update rewards full generations that (a) stay executable under our parser and (b) produce a one-step transition that enhances the topology-based progress signal. This aligns with the remote-control scenario: what ultimately matters is generating a valid key at every step and making steady progress along the latent UI graph, while still permitting diverse natural-language rationales during training.
\section{Comparison of TV UI Layouts}
\label{appsec:Comparison of TV UI Layouts}
\begin{figure*}[h]
    \centering
    \includegraphics[width=\textwidth]{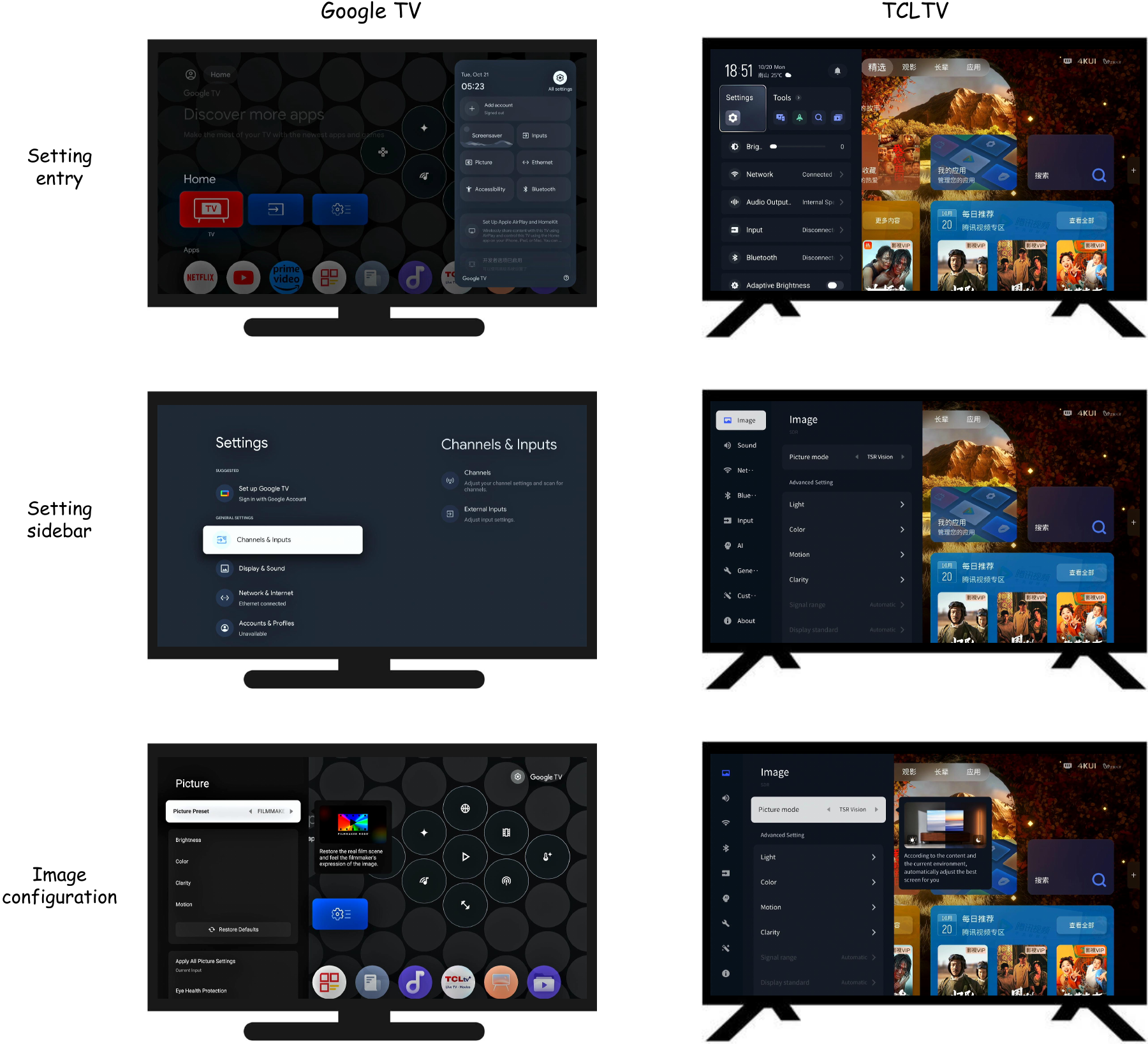}
    \caption{Comparison of UI style between two different TV models: Google TV and TCL TV.}
    \label{fig:Comparison of UI}
\end{figure*}

Figure~\ref{fig:Comparison of UI} provides a qualitative comparison of the UI layouts of Google TV and TCL TV. The two interfaces differ notably in icon appearance, overall layout aesthetics, and menu structure. This cross-platform variation naturally leads to a distribution shift in UI states, making it a suitable scenario for evaluating out-of-domain generalization in TV navigation.

\section{More experiments}
\label{appx: more_exp}

\paragraph{Effect of Different Rationale Types.}
As described in Sec.~\ref{sec:stage1}, we employ three types of step-level rationales during Stage I training. Table~\ref{tab:cot_ablation} reports an ablation study on text-based navigation tasks in TVWorld-N.
Removing rationale supervision yields the lowest success rate of $36.8$ (experiment (1)). Adding Geodesic Guidance alone improves performance to $42.4$ (experiment (2)), while further incorporating Detour Reflection or Stagnation Escape leads to consistent gains (experiments (3)–(4)). The best performance is achieved when all three rationale types are combined (experiment (5), $46.8$).
These results indicate that the three rationale types are complementary, with stagnation handling playing a particularly important role in remote-control TV navigation.

\begin{table}[h]
\centering
\caption{Ablation of Stage I Rationale Types for Text-Based Instructions on TVWorld-N. GG, DR, and SE denote Geodesic Guidance, Detour Reflection, and Stagnation Escape, respectively.}
\scalebox{1}{
\begin{tabular}{l |c | c}
\toprule
 & Rationale Types & SR  \\
\midrule
(1) & -   & 36.8 \\
(2) & GG  & 42.4 \\
(3) & GG \& DR & 42.8 \\
(4) & GG  \& SE & 46.4 \\
(5) & GG  \& DR \& SE  & \textbf{46.8} \\
\bottomrule
\end{tabular}}
\label{tab:cot_ablation}
\end{table}

\paragraph{Effect of Different Distance Metrics.}
We perform an ablation over various distance metrics in Stage~II to examine their influence on training, as reported in Table~\ref{tab:dist_ablation}. Shortest-path, Soft Shortest-Walk, and Hitting Time yield comparable SR, with Hitting Time performing marginally better, while Personalized PageRank (PPR) trails substantially. We attribute Hitting Time's modest advantage to its definition as the expected first-arrival time under random walks, which yields a more globally informative topological signal than metrics based solely on shortest paths. By contrast, PPR incorporates a restart mechanism that effectively assumes a certain probability of ``teleportation'' back to the starting point, a behavior that does not align with TV UI interaction patterns (e.g., after pressing \texttt{HOME} from a deep page, returning to the same deep state is often non-trivial); as a result, the induced distance signal conflicts with the {TV interaction logic}, leading to a pronounced drop in performance.

\begin{table}[h]
\small
\centering
\caption{Ablation of the distance metric in the Stage II on TVWorld-N.}
\scalebox{1}{
\begin{tabular}{l | cc}
\toprule
\multirow{2}{*}{Distance Metric} & \multicolumn{2}{c}{SR} \\
& text-based & vision-based \\
\midrule
Shortest-path & 66.4 & 67.2 \\
Soft Shortest-Walk & 65.6 & 68.4 \\
Personalized PageRank & 60.8 & 59.2 \\
Hitting Time & \textbf{67.2} & \textbf{68.8} \\

\bottomrule
\end{tabular}}
\label{tab:dist_ablation}
\end{table}

\paragraph{Effect of Topology-Aware Reward Design.}
As introduced in Sec.~\ref{sec:Topology Augmented RL}, we employ trace-specific reward designs in Stage II, assigning different reward functions to different trace types. The detailed formulations are provided in Appendix~\ref{app:trace-reward}. As a baseline, we adopt the reward design used for Geodesic Guidance traces as a \emph{standard reward}, denoted as $R_{\text{std}}(u_t, a; g)$, which assigns rewards solely based on changes in the distance to the goal:
\begin{equation}
\label{eq:standard-reward}
{\small
R_{\text{std}}(u_{t},a;g)=
\begin{cases}
1, & d(u',g) < d(u_{t},g),\\
0.2, & d(u',g) = d(u_{t},g),\\
0, & d(u',g) > d(u_{t},g),
\end{cases}
}
\end{equation}
Table~\ref{tab:reward_ablation} compares the standard reward with our proposed topology-aware reward. Across both text- and vision-based instructions, topology-aware rewards consistently yield higher success rates. This result indicates that incorporating trace-specific topology signals provides more fine-grained reward guidance, enabling the agent to learn stronger topology-aware navigation behaviors.

\begin{figure}[h]
    \centering
    \includegraphics[width=.45\textwidth]{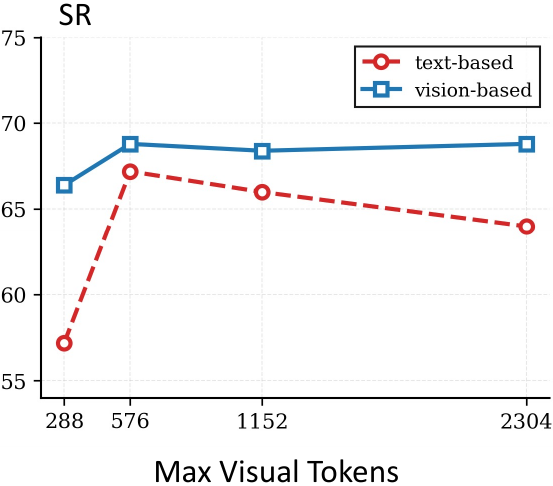}
    \caption{Model performance on TVWorld-N under different visual token budgets.}
    \label{fig:max_visual_token}
\end{figure}

\begin{table}[h]
\small
\centering
\caption{Ablation of the reward design in the Stage II on TVWorld-N.}
\scalebox{1}{
\begin{tabular}{l | cc}
\toprule
\multirow{2}{*}{Reward design} & \multicolumn{2}{c}{SR} \\
& text-based & vision-based \\
\midrule
standard reward & 64.0 & 67.6 \\
topology-aware reward & \textbf{67.2} & \textbf{68.8} \\

\bottomrule
\end{tabular}}
\label{tab:reward_ablation}
\end{table}
\paragraph{Effect of the Per-Image Visual Token Limit.}
In our default setting, each input image is resized to $1024\times576$, corresponding to $576$ visual tokens after processing by our model. Fig.~\ref{fig:max_visual_token} reports model performance under different per-image visual token limits. Increasing the token limit from 288 to 576 yields a substantial performance gain for both text-based and vision-based instructions, indicating that sufficient visual capacity is crucial for capturing salient UI details. Beyond this point, further increasing the token limit brings slight degradation, suggesting diminishing returns and potential noise introduced by overly fine-grained visual representations.

\begin{figure}[h]
    \centering
    \includegraphics[width=.45\textwidth]{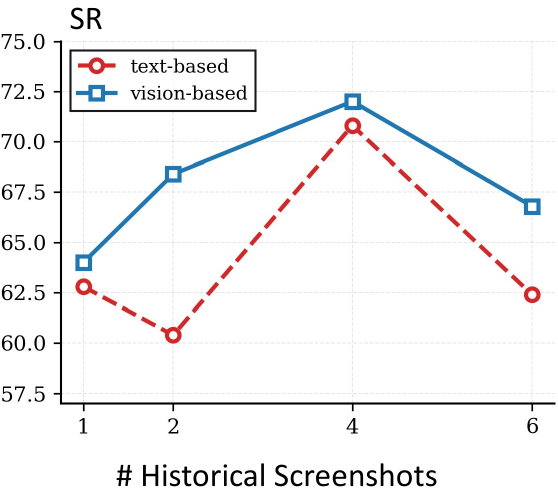}
    \caption{Model performance on TVWorld-N under different numbers of historical screenshots.}
    \label{fig:his_num}
\end{figure}

\paragraph{Effect of the Number of Historical Screenshots.}
TV navigation inherently involves long-horizon interactions, while screenshots introduce a substantial number of visual tokens, making it impractical to retain all historical screenshots as model input. In our default setting, we retain $4$ historical screenshots. Fig.~\ref{fig:his_num} illustrates the effect of varying the number of historical screenshots on model performance. We observe that using $4$ historical screenshots yields the best performance, while both increasing and decreasing this number lead to performance degradation.

\paragraph{Detailed Results on TVWorld-N.}
Table~\ref{tab:all_result} reports the detailed results of TVTheseus on TVWorld-N.

\begin{table*}[t!]
\centering
\caption{Detailed results of $3$ independent runs of TVTheseus on TVWorld-N.}
\scalebox{1}{
\begin{tabular}{l c ccccc c}
\toprule
Setting & Instr Type & Config & Display & Audio & Apps & Channels & Overall \\
\midrule

\multirow{2}{*}{Turn 1}
& text-based & 36.0 & 64.0 & 88.0 & 76.0 & 72.0 & 67.2 \\
& vision-based & 64.0 & 54.0 & 64.0 & 84.0 & 78.0 & 68.8 \\
\cmidrule(lr){1-8}
\multirow{2}{*}{Turn 2}
  & text-based & 40.0 & 52.0 & 88.0 & 68.0 & 72.0 & 64.0 \\
  & vision-based & 58.0 & 64.0 & 48.0 & 82.0 & 84.0 & 67.2 \\
\cmidrule(lr){1-8}
\multirow{2}{*}{Turn 3}
  & text-based & 48.0 & 72.0 & 82.0 & 68.0 & 84.0 & 70.8 \\
  & vision-based & 66.0 & 64.0 & 62.0 & 82.0 & 86.0 & 72.0 \\
\cmidrule(lr){1-8}
\multirow{2}{*}{Avg.}
  & text-based & 41.3 & 62.7 & 86.0 & 70.7 & 76.0 & 67.3 \\
  & vision-based & 62.7 & 60.7 & 58.0 & 82.7 & 82.7 & 69.3 \\
\bottomrule
\end{tabular}
}
\label{tab:all_result}
\end{table*}

\section{Case study}\label{appx: case_study}
\begin{figure*}[h]
    \centering
    \includegraphics[width=0.95\textwidth]{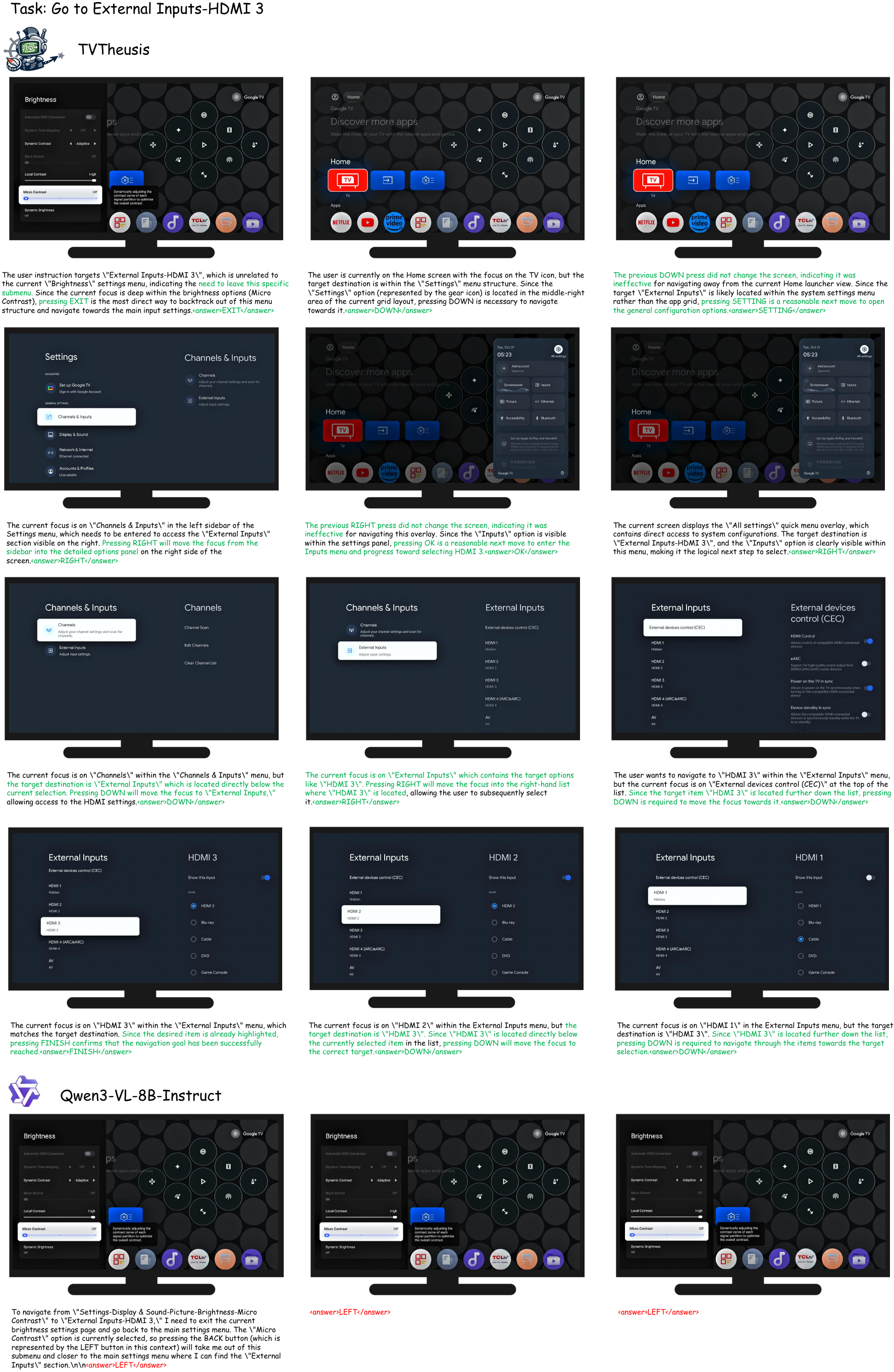}
    \caption{Case study comparing TV navigation trajectories before (Qwen3-VL-8B-Instruct) and after topology-aware training (TVTheseus).}
    \label{fig:casestudy1}
\end{figure*}

Figure~\ref{fig:casestudy1} shows a case study on focus-based TV navigation for the instruction “Go to External Inputs–HDMI 3,” comparing TVTheseus (topology-aware trained) with Qwen3-VL-8B-Instruct (untrained baseline) from the same initial UI state. TVTheseus plans a coherent sequence that navigates the settings hierarchy into External Inputs, shifts focus step-by-step to HDMI 3, and ends with \texttt{FINISH}; when a key press causes no state change, it adapts by trying alternative actions instead of repeating the ineffective one, demonstrating reliable topology-aware planning. In contrast, the untrained model repeatedly issues actions with no transitions, stays near the initial state, and shows limited understanding of focus-based UI dynamics and global planning.

\section{Training Data Example}
\label{sec:Training Data Example}
\begin{figure*}[h]
    \centering
    \includegraphics[width=\textwidth]{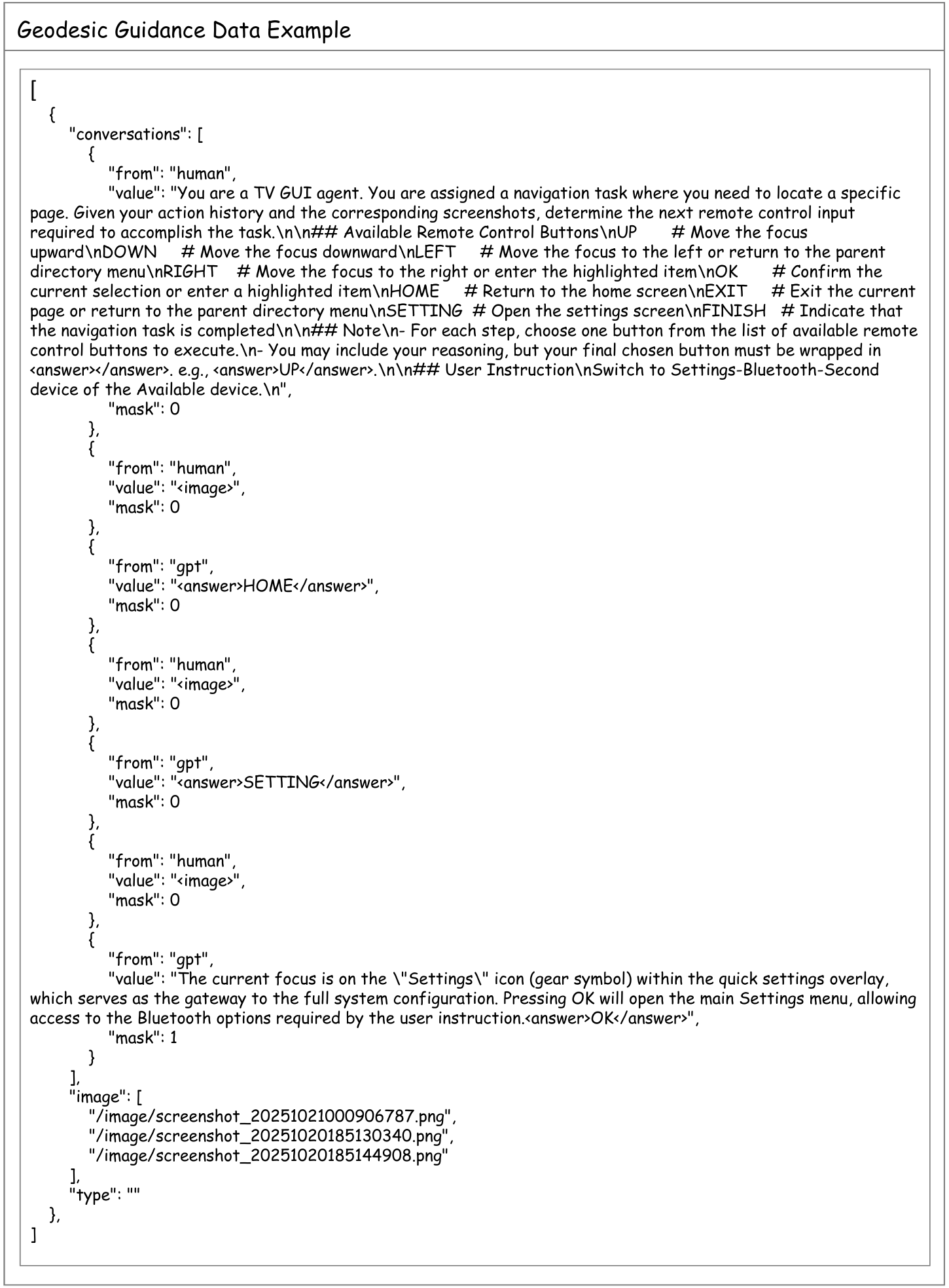}
    \caption{Example training data for Geodesic Guidance Traces.}
    \label{fig:geodataexample}
\end{figure*}

\begin{figure*}[h]
    \centering
    \includegraphics[width=0.9\textwidth]{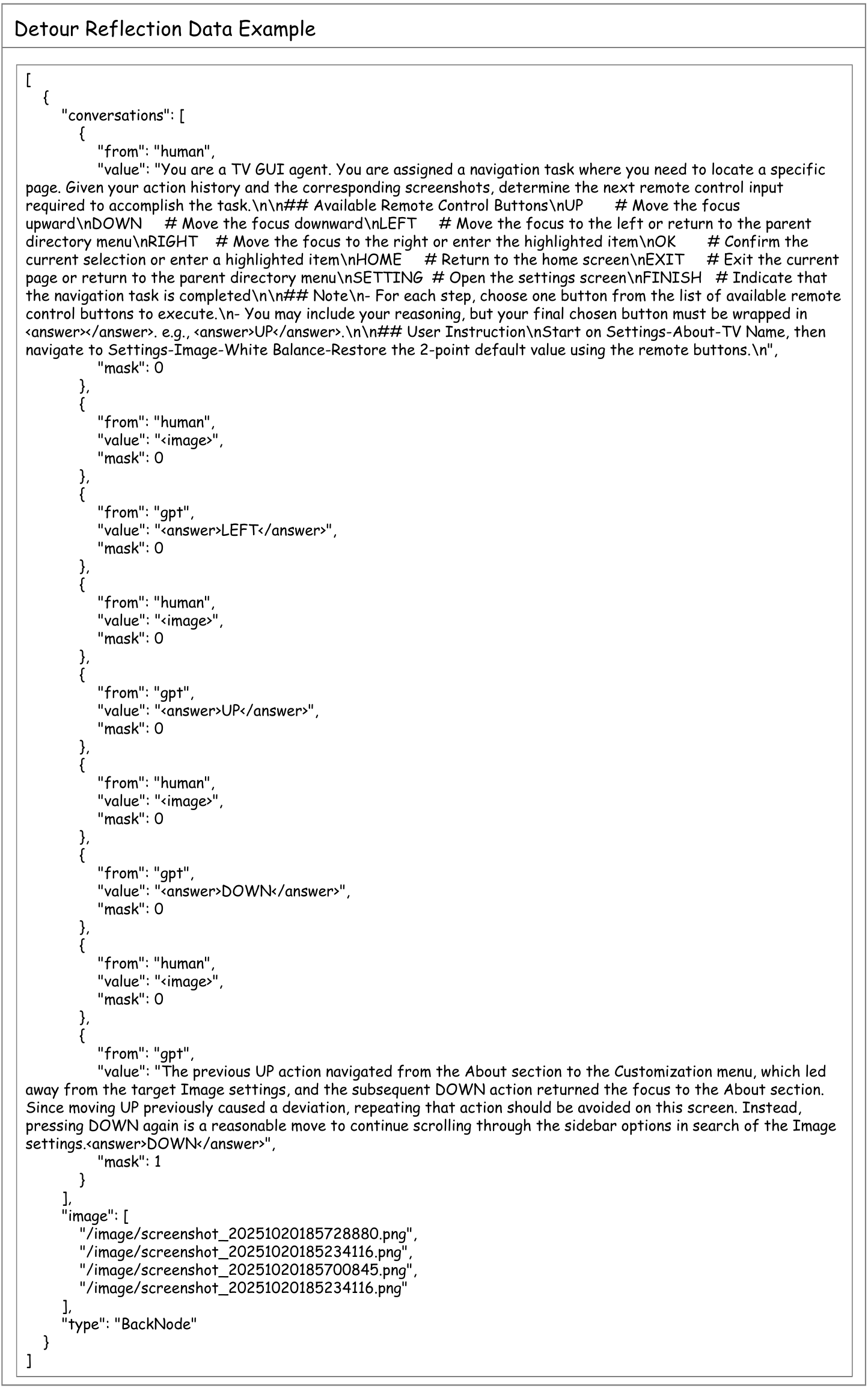}
    \caption{Example training data for Detour Reflection Traces.}
    \label{fig:detourdataexample}
\end{figure*}

\begin{figure*}[h]
    \centering
    \includegraphics[width=\textwidth]{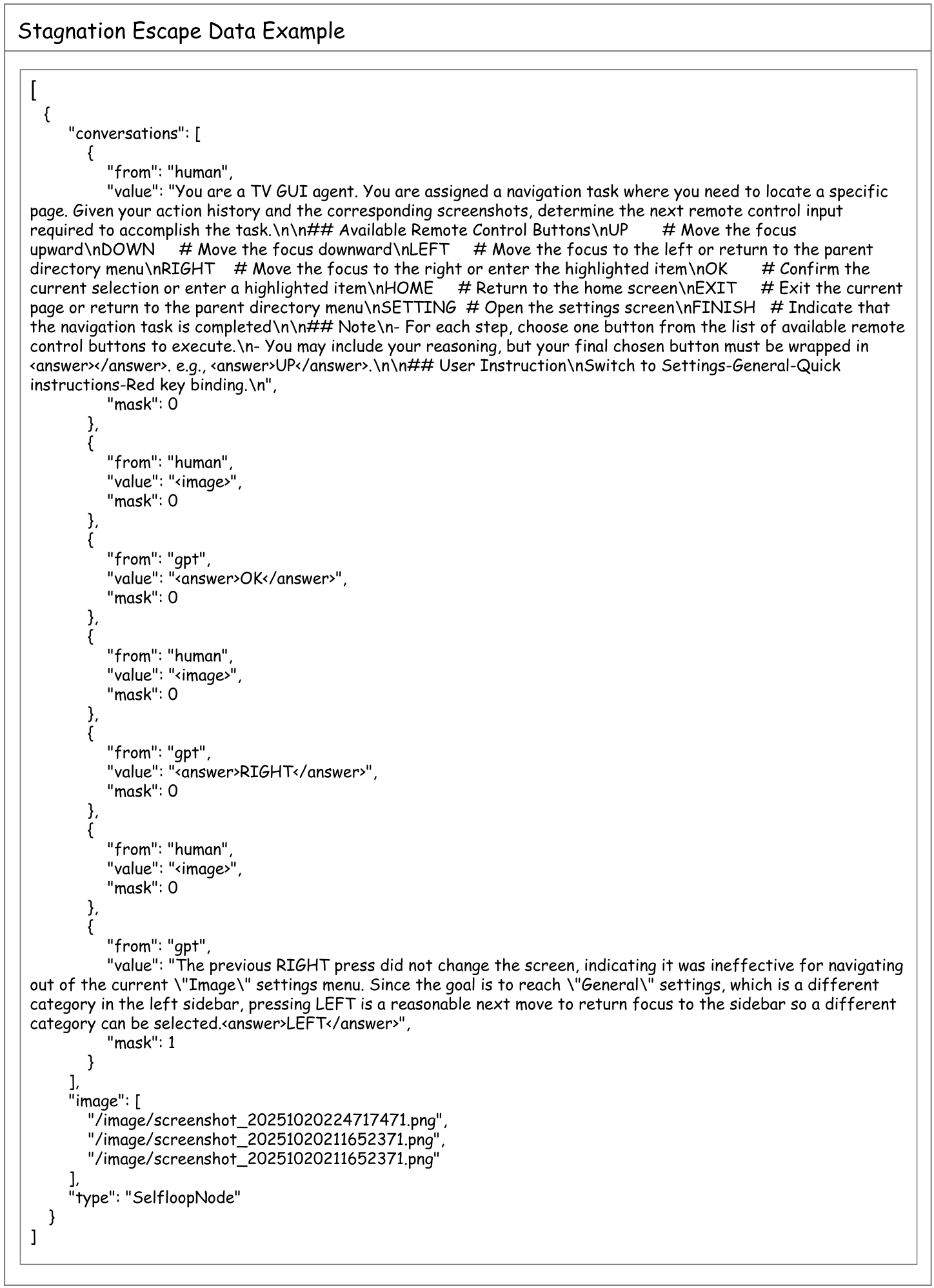}
    \caption{Example training data for Stagnation Escape Traces.}
    \label{fig:stagdataexample}
\end{figure*}

We present three categories of topology-priming SFT training data: {Geodesic Guidance} in Figure~\ref{fig:geodataexample}, {Detour Reflection} in Figure~\ref{fig:detourdataexample}, and {Stagnation Escape} in Figure~\ref{fig:stagdataexample}. Each instance contains {chain-of-thought} reasoning, records of past actions, and both historical and current page screenshots.

\section{Responsible NLP Research Considerations}
\subsection{Potential Risks}
If deployed on real devices, agents trained with TVWorld could be exploited to automatically access and change privacy- or account-related settings (such as permissions, parental controls, or password options), potentially causing privacy or security harms. We partly reduce this risk by masking a limited set of sensitive entry points during graph construction and release, and we advise using access controls and explicit user confirmation for any deployment on real devices.
\subsection{Intended Use \& Artifact Use}
TVWorld (and TVWorld-N/TVWorld-G) and TVTheseus are intended for research on focus-based remote-control TV navigation, including controlled training and benchmarking of agents in an offline, replayable environment. They are not intended for deployment on unauthorized control of devices, or attempts to access restricted system functions. We use existing models and tools strictly in accordance with their intended research/benchmarking usage and applicable terms; we do not provide the system with any personal user data. We recommend that any derivatives of the released assets remain limited to research contexts.
\subsection{AI Assistants Elaboration}
In this work, we employed AI assistants strictly as supporting tools for tasks including grammar correction, language refinement, and logo image generation. The authors thoroughly evaluated and revised all outputs provided by these tools and retain complete responsibility for the accuracy, integrity, and content of the final manuscript.
\subsection{Ethics and Reproducibility Statement}
We study offline, replayable TV-navigation agents using TVWorld/TVTheseus, constructing static graphs and screenshots to support reproducible evaluation.
We have checked that the collected/used data do not contain any personally identifiable information, including identifiable personal names, and do not include any private or sensitive user information.
All external datasets, models, and tools used in this work are properly cited and employed in full compliance with their licenses, terms, and intended-use policies.
As such, we do not anticipate potential ethical risks arising from the dataset or experimental protocol.
To further promote transparency and reproducibility, we release all code, models, datasets, and related resources on public platforms such as GitHub and Hugging Face under the CC BY 4.0 and Apache-2.0 licenses.


\end{document}